\documentclass{article} 
\usepackage{iclr2027_conference,times}
\usepackage{graphicx}
\usepackage{booktabs}
\usepackage{multirow}
\usepackage{amsmath}
\usepackage{amssymb}
\usepackage{amsfonts}
\usepackage{mathtools}
\usepackage{amsmath,amssymb,algorithm,algpseudocode,float,placeins}
\usepackage{hyperref}
\usepackage{cleveref}
\usepackage{booktabs}
\usepackage{tabularx}
\usepackage{microtype}
\usepackage{wrapfig}
\usepackage{booktabs,wrapfig,needspace}
\usepackage{capt-of}

\usepackage{amsmath,amsfonts,bm}

\def\eqref#1{equation~\ref{#1}}

\def\1{\bm{1}}

\DeclareMathAlphabet{\mathsfit}{\encodingdefault}{\sfdefault}{m}{sl}
\SetMathAlphabet{\mathsfit}{bold}{\encodingdefault}{\sfdefault}{bx}{n}

\usepackage{hyperref}
\usepackage{url}

\title{Interaction Latents as Actions for Contact-Rich Manipulation}

\author{
\begin{tabular}{@{}l@{}}
\textbf{
Yuhang Zheng$^{1,*}$ \quad
Yujie Zang$^{1,*}$ \quad
Yuchen Liu$^{2,4,*}$ \quad
Yupeng Zheng$^{2,3,5,\dagger}$
} \\
\textbf{
Zhennan Jiang$^{5}$ \quad
Zebin Xing$^{3,5}$ \quad
Shuai Tian$^{3,5}$ \quad
Songen Gu$^{3,6}$
} \\
\textbf{
Jingxiang Guo$^{1}$ \quad
Chen Gao$^{1,\dagger}$ \quad
Zhongpu Xia$^{2,+}$ \quad
Shuicheng Yan$^{1,\dagger}$
} \\[6pt]
{\normalfont\small
$^{1}$ NUS} \quad
{\normalfont\small
$^{2}$ Anyverse Robotics \quad
$^{3}$ Open-Phoenix Team} \quad
{\normalfont\small
$^{4}$ BIT} \quad
{\normalfont\small
$^{5}$ UCAS} \quad
{\normalfont\small
$^{6}$ FDU} \\[4pt]
{\normalfont\footnotesize
$^{*}$ Equal contribution \quad
$^{+}$ Project lead \quad
$^{\dagger}$ Corresponding authors}
\end{tabular}
}

\iclrfinalcopy 
\begin{document}

\maketitle
\fancyhead{} 

\begin{abstract}
Contact-rich manipulation requires robots to coordinate not only where to
move, but also how contact and force should evolve throughout an interaction.
Yet most robot policies represent actions primarily as
motion trajectories, using tactile signals as inputs or auxiliary
prediction targets without explicitly encoding contact dynamics in
the action representation.
We propose \textbf{ILA (Interaction Latents as Actions)}, which uses decodable interaction latents that jointly capture the evolution of motion, contact, and force application as policy action units, providing a shared representation for action generation, evaluation, and offline improvement.
We learn these latents through visually conditioned joint action-tactile reconstruction with a time-conditioned implicit decoder.
Using them as flow-matching generation targets supports post-training of pretrained vision-language-action models (VLAs) and world action models (WAMs), as well as training flow-matching policies from scratch.
A shared decoder maps generated latents to executable actions.
For offline improvement, we learn a tactile-conditioned expert reference model from interaction latents encoded from expert demonstrations.
The resulting expert-consistency rewards guide weighted learning from successful rollouts, while observation-compatible expert references guide correction of low-reward predictions from failed rollouts.
Across three policy families, ILA improves success rates averaged over four real-world tasks by 21.0–36.5 percentage points relative to tactile-conditioned direct-action baselines. Ablations highlight the benefit of encoding contact evolution in action targets beyond tactile conditioning, while the same latent space supports expert-guided offline improvement. Project website: \url{https://mrsecant.github.io/ILA/}.

\end{abstract}

\section{Introduction}
\label{sec:introduction}

Contact-rich manipulation requires coordinating motion and force application with object properties, motion states, and contact conditions.
Motion trajectories alone may not capture these interactions: similar movements can produce different contact patterns, while the same task may require different motion--force combinations~\citep{liu2025forcemimic,Wirnshofer2020ControllingCM}.
A key challenge is to generate motion while accounting for contact states, force magnitudes, and their temporal evolution.

Prior work incorporates tactile and force signals into policy learning~\citep{Dong2026ForceDeltaVLADF, bi2026vla, he2025foar, huang20243d}.
Visuo-tactile policies such as RDP~\citep{xue2025reactive} and T-Rex~\citep{Niu2026TRexTD} use high-frequency tactile feedback to adapt actions to contact changes.
World action models (WAMs), such as VT-WAM~\citep{tian2026vt} and N0-TWAM~\citep{team2026n_0}, jointly predict future vision, touch, and actions, using anticipated contact evolution to guide action generation.
However, tactile conditioning or parallel prediction does not require generated action units to support both motion and tactile reconstruction.
Contact and force information may thus be lost from the action representation, potentially limiting motion-contact coordination.
This motivates a different question: \emph{can motion and physical interaction
be represented jointly as the action itself?}

We introduce \textbf{ILA (Interaction Latents as Actions)}, which represents
the joint evolution of motion, contact, and force application as a decodable
policy action unit.
ILA learns these interaction latents through visually conditioned joint
action--tactile reconstruction, and policies generate them instead of motor
trajectories directly.
A shared decoder maps the generated latents to executable actions.
Importantly, the same interaction space also supports expert-consistency
evaluation and reference-guided policy correction, providing a common
representation for action generation, evaluation, and offline policy improvement.

\begin{figure}[t]
    \centering
    \includegraphics[width=0.92\linewidth]{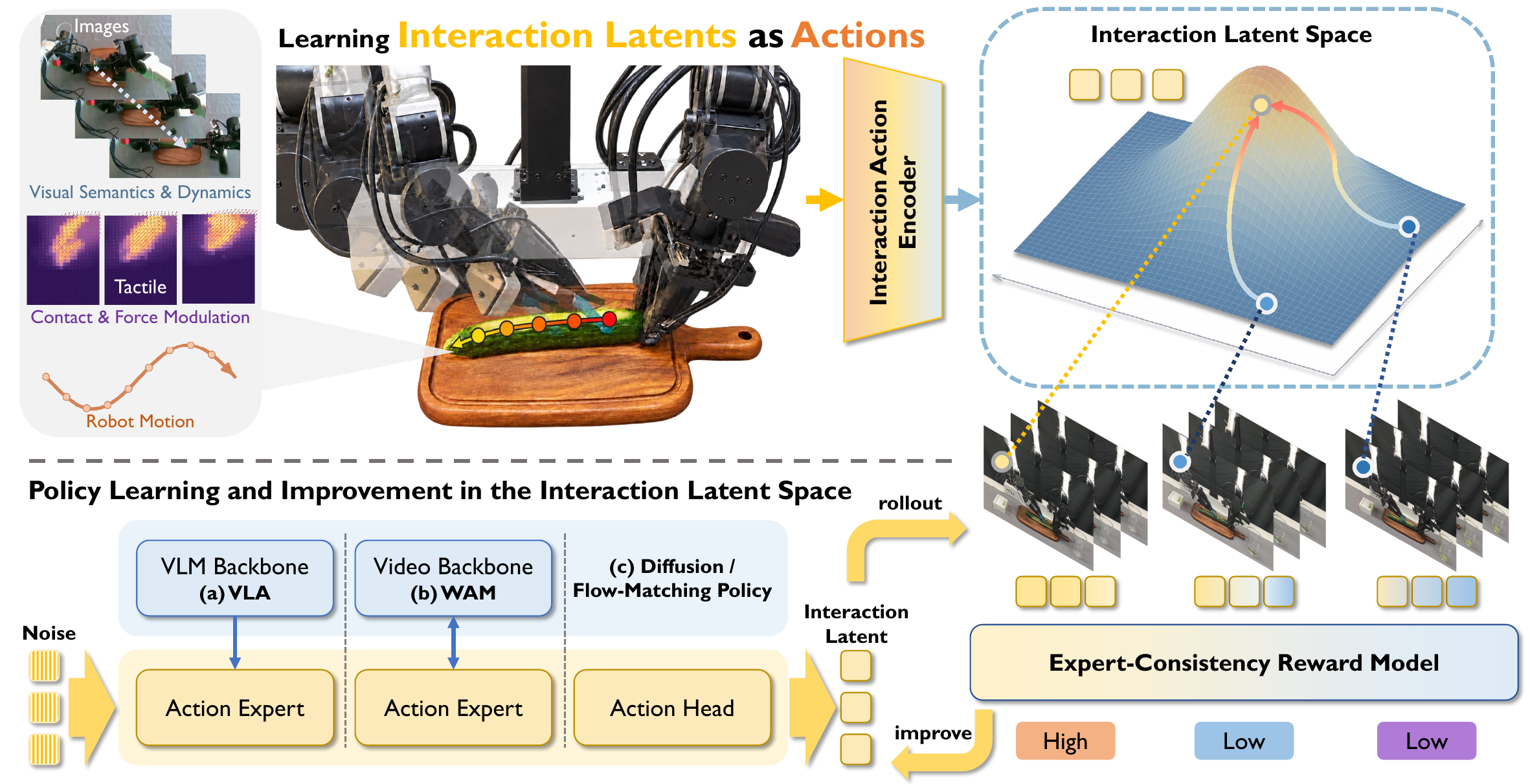}
    \vspace{-2mm}
    \caption{
    \textbf{Interaction Latents as Actions (ILA).}
    A shared latent action jointly represents motion and tactile evolution over temporally extended interactions and decodes into executable actions.
    Policies across architectures generate these latents, while expert-consistency rewards and compatible references guide offline improvement, unifying action generation, evaluation, and correction.
    }
    \vspace{-4mm}
    \label{fig:teaser}
\end{figure}

Specifically, we encode action and tactile sequences into a shared bottleneck.
A time-conditioned implicit decoder reconstructs actions, tactile deformation fields, and aggregate deformation, encouraging the representation to preserve motion, local contact, overall force application, and their temporal relationships.
We use these latents as flow-matching targets to post-train vision-language-action models (VLAs) and WAMs and to train flow-matching policies from scratch, retaining each architecture's native conditioning mechanism.
For offline improvement, expert demonstrations are encoded into the same
latent space to define a tactile-conditioned expert reference model.
Expert-consistency rewards then weight learning from successful policy
rollouts, while observation-compatible expert references provide correction
targets for low-reward predictions from failed rollouts.
Thus, policy generation, interaction evaluation, and correction all operate
on the same interaction representation.

We evaluate ILA on four real-world tasks: cucumber peeling, fried egg flipping, celery cutting, and plate wiping.
Comparisons with direct motion generation, tactile conditioning, and joint action—tactile prediction isolate the role of interaction action units.
Modality ablations and representation analyses examine how motion and contact information are retained and how they contribute to policy learning.
We also evaluate expert-consistency rewards and reference-guided correction.
ILA improves average success rates by 21.0–36.5\% over tactile-conditioned direct-action baselines across policy families, supporting joint motion–contact representations as effective action-generation targets. Refining FM-ILA in the same latent space reduces force-proxy error by 75\% on cucumber peeling while increasing success, demonstrating the value of interaction latents for both action generation and offline improvement.

In summary, our contributions are:
\begin{itemize}
    \item \textbf{Interaction latents as policy actions.}
    We introduce a decodable action representation that jointly captures
    motion, contact, and force evolution, and use it as a common generation
    target across VLA, WAM, and flow-matching policies.

    \item \textbf{Interaction-space evaluation and offline improvement.}
    We use expert demonstrations to define tactile-conditioned
    expert-consistency rewards in the same latent space, combining
    reward-weighted learning from successful rollouts with
    observation-compatible expert correction of failed interactions.

    \item \textbf{Real-world evaluation across policy architectures.}
    Experiments on four contact-rich tasks show consistent gains from ILA
    across three policy families, together with motion--contact representation
    analyses and expert-guided offline improvement.
\end{itemize}

\section{related work}
\subsection{Tactile Representations and Tactile-Guided Manipulation}
\vspace{-1mm}
Tactile representation learning captures transferable information about object properties and contact states.
Sparsh~\citep{higuera2024sparsh} and T3~\citep{zhao2024transferable} promote transfer through self-supervised pretraining and cross-sensor learning, respectively.
UniTouch~\citep{yang2024binding} and AnyTouch~\citep{Feng2025AnyTouchLU} align touch with vision, with the latter also unifying static and dynamic tactile representations.
AnyTouch 2~\citep{feng2026anytouch}, TaF-VLA~\citep{huang2026taf}, and exUMI~\citep{Xu2025exUMIER} extend this line toward temporal contact dynamics through action-related deformation modeling, tactile-force alignment, and action-aware tactile prediction.
In manipulation, tactile information supports both feedback control and future prediction.
RDP~\citep{xue2025reactive} and T-Rex~\citep{Niu2026TRexTD} use high-frequency tactile feedback for rapid action refinement~\citep{Zhou2026TouchWorldAP}.
Beyond responding to current contact, TacForeSight~\citep{zang2026tacforesight} anticipates contact through force-conditioned tactile prediction, while OmniVTA~\citep{zheng2026omnivta} integrates visuo-tactile prediction with action generation and feedback correction~\citep{Zhang2026ImaginingTS}.
Dream-Tac~\citep{lou2026dream}, VT-WAM~\citep{tian2026vt}, ME-Dex~\citep{zhang2026me} and TacWAM~\citep{jin2026tacwam} further unify tactile and action prediction within world action models~\citep{Zhang2026ContactWorldWR,Qin2026DexTouchWMLA}.
While these methods use touch to support action generation, we make motion-contact coordination part of the action representation itself.

\subsection{Reward Learning for Manipulation}
\vspace{-1mm}
Reward learning turns demonstrations and interaction experience into training signals, reducing reliance on task-specific reward engineering.
T-REX~\citep{brown2019extrapolating} learns rewards from ranked demonstrations.
Progress-based methods derive rewards from learned observation representations~\citep{zakka2022xirl,ma2022vip,biza2025robot,Tang2025ReLAMLA}.
GCR combines temporal value learning with success-failure discrimination, while XIRL and VIP relate goal proximity to task progress.
SRPO~\citep{fei2026srpo} measures progress relative to successful policy rollouts in a pretrained world-model latent space.
SARM~\citep{chen2025sarm} learns stage-aware progress rewards to filter and reweight demonstrations.
Generative approaches derive rewards from video likelihoods~\citep{escontrela2023video}, conditional video entropy~\citep{Huang2023DiffusionReward}, or latent-transition agreement~\citep{zhang2026dream2reward}.
DIFO~\citep{huang2024diffusion} uses diffusion-based discrimination of state transitions, while DRAIL~\citep{lai2024drail} and FA-OPD~\citep{wan2026adversarial} derive adversarial state-action rewards from diffusion errors and class-conditioned flow-matching residuals, respectively.
For contact-rich manipulation, HapticVLA~\citep{gubernatorov2026hapticvla} uses safety-aware tactile rewards for reward-weighted flow matching.
Our method evaluates policy-predicted interaction chunks in a joint action-tactile latent space, using tactile-conditioned expert-consistency rewards and observation-compatible reference corrections to guide offline policy improvement.
\section{Method}
\label{sec:method}

We propose \textbf{ILA (Interaction Latents as Actions)}, which jointly represents motion trajectories, contact evolution, and force application as a decodable \emph{interaction latent}: a common action unit for policy generation, evaluation, and improvement.
We learn this representation through visually conditioned joint action--tactile reconstruction, enabling different policy architectures to generate future interactions in the same action space.
Within this space, we further learn a tactile-conditioned expert reference distribution and use chunk-level expert-consistency rewards and compatible reference interactions to guide offline policy improvement.

\subsection{Learning Joint Interaction Action Representations}
\label{sec:interaction_representation}

Our goal is to learn an action representation that describes how the robot moves, makes contact, and applies force.
To this end, we jointly encode temporally aligned action and tactile sequences under visual conditioning into an interaction latent~\citep{Lee2018MakingSO, Lee2024BehaviorGW}.

\paragraph{Multimodal interaction encoding.}
For a demonstration chunk starting at time $t$, let $A_t$ and $S_t$ denote the aligned $H$-step sequences of actions and three-dimensional tactile deformation fields, and $I_t^{\mathrm{hist}}$ the multi-view image history available at the chunk's start.
The \emph{interaction encoder} encodes actions and touch conditioned on this visual history:
\begin{equation}
z_t^{\mathrm{enc}}=E_{\phi}(A_t,S_t;V_t)\in\mathbb R^{K\times d_z},
\qquad V_t=E_{\mathrm{vis}}(I_t^{\mathrm{hist}}),
\label{eq:interaction_encoding}
\end{equation}
where $K<H$ is the number of temporally compressed tokens and $d_z$ is the feature dimension.
The visual branch provides scene semantics and motion context for joint motion--contact encoding.
A frozen DINOv2~\citep{Oquab2023DINOv2LR} encoder extracts patch features from multiple frames, with current-frame features representing scene semantics.
Local spatiotemporal attention and adjacent-frame feature differences capture implicit and explicit motion cues, respectively.
These motion features are fused and combined with current-frame semantics at the patch level to form semantics-and-dynamics (S\&D) tokens.
Action and tactile sequences are separately encoded into temporally aligned tokens, coupled through joint self-attention, and conditioned on visual context through cross-attention before temporal compression into $z_t^{\mathrm{enc}}$~\citep{Vaswani2017AttentionIA}.
The decoder reconstructs both modalities from the same joint representation, without a separate visual input.
This asymmetry routes the contribution of vision to reconstruction through the shared bottleneck, encouraging interaction-relevant context to enter the action representation itself.

\begin{figure}[]
    \centering
    \includegraphics[width=0.95\linewidth]{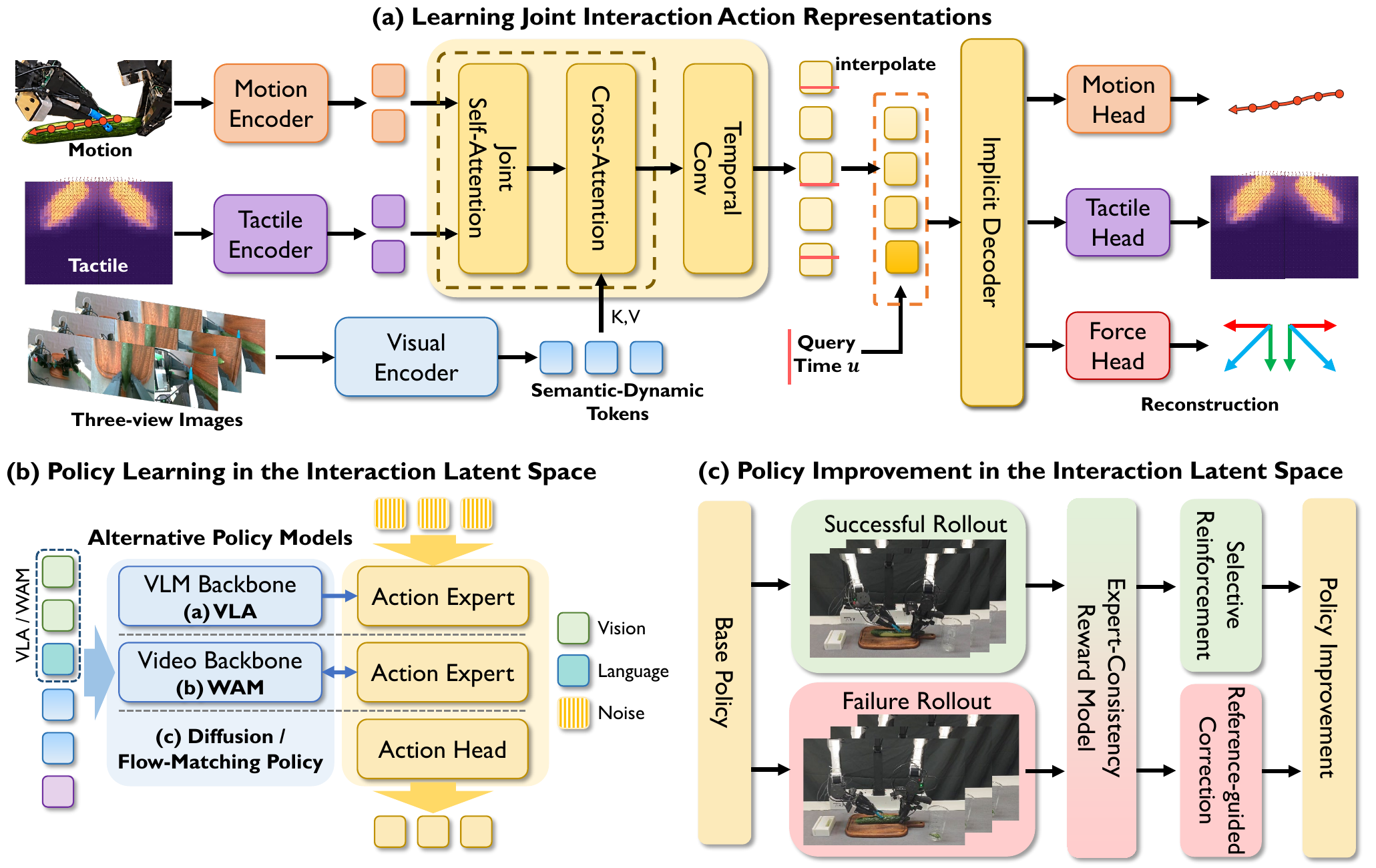}
    \vspace{-2mm}
    \caption{
    \textbf{Overview of our proposed ILA for policy learning and offline improvement.}
    }
    \vspace{-2mm}
    \label{fig:pipeline}
\end{figure}

\paragraph{Implicit joint reconstruction.}
To capture how motion and contact evolve together, we use a time-conditioned implicit decoder that represents both as continuous-time functions conditioned on the same joint latent~\citep{Chen2020LearningCI}.
For a query time $u$ within the chunk, we linearly interpolate neighboring tokens in $z_t^{\mathrm{enc}}$ and fuse the interpolated feature with a time encoding.
Shared decoding features then predict the action, tactile deformation field, and aggregate deformation at that time.

During training, we augment the original $H$ observation times with $2H$ query times drawn by stratified random sampling.
Action and deformation targets at these additional times are obtained by linearly interpolating adjacent observed frames; aggregate deformation targets are computed from the interpolated fields.
We use the same decoder and supervision terms for joint reconstruction at the original sampling times and additional inter-frame query times.
We use three-dimensional deformation fields as proxies for distributed forces, and the spatial aggregates of their tangential components, termed \emph{aggregate deformation}, as proxies for overall tangential force application.
Pointwise reconstruction captures local contact structure, whereas overall force application also depends on how deformation accumulates spatially.
To explicitly preserve this information, a lightweight aggregate deformation head supervises the shared decoding features, while a separate loss on the reconstructed field's aggregate directly constrains the tactile reconstruction branch.

Let $\mathcal Q_{\mathrm{obs}}$ and $\mathcal Q_{\mathrm{int}}$ denote the observed and interpolated query sets.
The overall training objective is
\begin{equation}
\begin{aligned}
\mathcal L_{\mathrm{repr}}
&=\mathcal L_{\mathrm{rec}}(\mathcal Q_{\mathrm{obs}})
+\lambda_{\mathrm{int}}\mathcal L_{\mathrm{rec}}(\mathcal Q_{\mathrm{int}}),\\
\mathcal L_{\mathrm{rec}}(\mathcal Q)
&=\mathcal L_{\mathrm{act}}(\mathcal Q)
+\lambda_{\mathrm{tac}}\mathcal L_{\mathrm{tac}}(\mathcal Q)
+\lambda_{\mathrm{dir}}\mathcal L_{\mathrm{dir}}(\mathcal Q)
+\lambda_{\mathrm{map}}\mathcal L_{\mathrm{map}}(\mathcal Q).
\end{aligned}
\label{eq:interaction_reconstruction}
\end{equation}
Each term is averaged over its respective query set $\mathcal Q$.
$\mathcal L_{\mathrm{act}}$ and $\mathcal L_{\mathrm{tac}}$ are elementwise $L_1$ reconstruction losses on normalized actions and deformation fields, respectively.
$\mathcal L_{\mathrm{dir}}$ and $\mathcal L_{\mathrm{map}}$ supervise directly predicted aggregates and aggregates computed from reconstructed fields; each combines $L_1$ errors in vector components and magnitudes.
Deformations are aggregated in their original scale and rescaled by a fixed per-sensor RMS statistic from the training set before computing these losses.
Together, these objectives shape a single decodable interaction action representation through motion trajectories, local contact, force application, and their temporal evolution.

\subsection{Policy Learning in the Interaction Latent Space}
\label{sec:interaction_policy}

We use the joint interaction representation as a unified target for flow-matching action generation, allowing policies to learn future motion and contact patterns from observations.
This action space supports both post-training VLAs and WAMs and training flow-matching policies from scratch.

We freeze the Interaction Action Encoder, its visual branch, and the decoder, and encode demonstration action--tactile chunks offline to obtain supervision latents $z_t^{\mathrm{enc}}$.
Using their normalized values as generation targets, we learn the conditional policy $\pi_\theta(\cdot\mid o_t)$ through chunk-level flow matching, conditioned on current and past visual, tactile, and proprioceptive observations available at the chunk's start.
Each architecture incorporates these conditions through its native conditioning mechanism.
For VLAs, we retain native visual and language inputs and append semantics-and-dynamics tokens and tactile and proprioceptive history tokens to the VLM prefix, which conditions the action expert's generation of interaction latents.
For WAMs, we retain native video inputs and the video generation objective, combine the additional tokens with language into shared cross-attention conditions for the video and action branches, and replace the action branch's generation target with interaction latents.
For flow-matching policies trained from scratch, we directly condition a DiT on semantics-and-dynamics tokens and tactile and proprioceptive history tokens to generate latents in the same space~\citep{Lipman2022FlowMF}.

At inference, the predicted latent $\hat z_t$ is denormalized and decoded by the frozen action decoder on a fixed 30\,Hz time grid into joint target sequences.
Different architectures can thus generate actions through a joint motion--contact representation while preserving the original execution interface.
The same policy predictions also serve as the objects of subsequent chunk-level expert-consistency evaluation and offline policy improvement.

\subsection{Offline Policy Improvement in the Interaction Latent Space}
\label{sec:interaction_improvement}

We evaluate policy predictions against expert interactions and use
observation-compatible expert references to guide offline improvement in
the same latent space.

\paragraph{Expert-consistency rewards.}
We encode recorded action--tactile chunks from expert demonstrations with
the frozen interaction encoder, forming expert reference latents
$z_j^e \in \mathcal{D}_E$ in the policy's normalized coordinates.
These encoded expert interactions directly define the reference set and
the training targets for the reward model.
We train $v_\omega$ on these latents through flow matching.

To evaluate a policy prediction $\hat z_t$, we sample flow time $s$ from the reward-evaluation distribution $\nu$ and Gaussian noise $\epsilon$, and construct $z_t^{\mathrm{noise}}=(1-s)\epsilon+s\hat z_t$.
We define an expert-consistency energy and reward using the conditional flow-matching residual:
\begin{equation}
\begin{aligned}
\mathcal E_\omega(\hat z_t,c_t)
&=\mathbb E_{s\sim\nu,\,\epsilon}
\left[
\frac{1}{K d_z}
\left\|v_\omega(z_t^{\mathrm{noise}},s,c_t)
-(\hat z_t-\epsilon)\right\|_F^2
\right],\\
r_t&=-\mathcal E_\omega(\hat z_t,c_t).
\end{aligned}
\label{eq:expert_consistency_reward}
\end{equation}
Here, $c_t$ denotes the aligned tactile descriptor used for offline scoring. Lower energy indicates closer alignment with the conditional transport field learned from expert references.
This energy measures expert consistency rather than an explicit negative log-likelihood, providing a continuous reward for chunk weighting and reference-guided correction.

\paragraph{Selective reinforcement.}
We collect interaction chunks from successful rollouts into $\mathcal D^+$ and train on their recorded policy predictions through reward-weighted flow matching~\citep{Peng2019AdvantageWeightedRS,Nair2020AcceleratingOR}.
The weight $w(r)$ combines a uniform term with a reward-dependent term, emphasizing high-reward interactions while retaining learning from other successful samples.

\paragraph{Reference-guided correction.}
For low-reward chunks from failed rollouts, we retrieve high-reward expert references with compatible observations from $\mathcal D_E$ and construct latent correction targets under the original observation conditions.
Let $\mathcal P=\{(i,j)\}$ denote the resulting pairs, where $i$ indexes a chunk to be corrected and $j$ its expert reference.

Let $h_\theta$ denote the noise-free latent estimator of the same policy introduced in Section~\ref{sec:interaction_policy}, initialized from a frozen copy of the policy prior to improvement, $h^{\mathrm{ref}}$.
For a failed prediction $\hat z_i$, we construct $z_i^{\mathrm{noise}}=(1-s)\epsilon+s\hat z_i$ and obtain estimates from both models using this input and the original observation $o_i$.
Taking the frozen estimate $\hat z_i^{\mathrm{ref}}=h^{\mathrm{ref}}(z_i^{\mathrm{noise}},s\mid o_i)$ as the baseline, we construct a compatibility-adaptive correction target:
\begin{equation}
z_{ij}^{*}=(1-\alpha_{ij})\hat z_i^{\mathrm{ref}}+\alpha_{ij} z_j^e,
\qquad \alpha_{ij}=\exp(-d_{ij}/T_{\mathrm{comp}}).
\label{eq:reference_correction}
\end{equation}
Here, $d_{ij}\geq0$ measures observation discrepancy, and $T_{\mathrm{comp}}>0$ controls the locality of reference transfer.
More compatible observations yield targets closer to the expert reference; less compatible observations retain more of the frozen policy's estimate.
We learn this correction from the noisy latent derived from the failed prediction, keeping the pre-improvement policy frozen.

Reward weighting and reference correction jointly define the offline policy improvement objective:
\begin{equation}
\begin{aligned}
\mathcal L_{\mathrm{policy}}(\theta)
={}&\mathbb E_{(o_t,\hat z_t,r_t)\sim\mathcal D^+}
\left[w(r_t)\,\ell_{\mathrm{FM}}(\theta;o_t,\hat z_t)\right]\\
&+\lambda_{\mathrm{corr}}
\mathbb E_{(i,j)\sim\mathcal P,\,s,\epsilon}
\left[
\lambda(s)
\left\|h_\theta(z_i^{\mathrm{noise}},s\mid o_i)
-\operatorname{sg}(z_{ij}^{*})\right\|_F^2
\right].
\end{aligned}
\label{eq:offline_policy_improvement}
\end{equation}
Here, $\ell_{\mathrm{FM}}$ is the policy's chunk-level flow-matching loss, $\lambda(s)$ weights noise levels, $\operatorname{sg}$ stops gradients through the correction target, and $\lambda_{\mathrm{corr}}$ balances the two updates.
Reference set construction, reward weighting, and sampling details are provided in the supplementary material.

The joint interaction representation thus serves as both the action content generated by the policy and the object of expert-consistency evaluation and reference-guided correction, carrying motion--contact coordination through action learning and offline policy improvement.

\begin{figure}[t]
    \centering
    \includegraphics[width=0.9\linewidth]{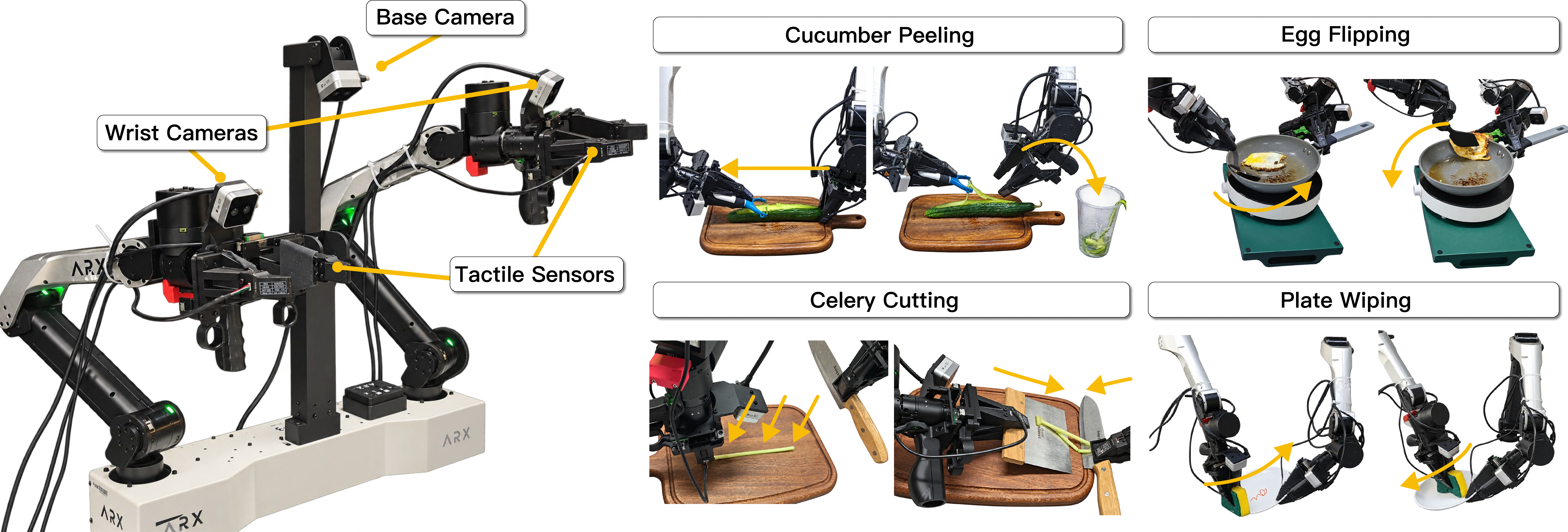}
    \vspace{-2mm}
    \caption{
    \textbf{Experimental setup and contact-rich manipulation tasks.}
    Left: dual-arm robotic platform equipped with wrist and base cameras and fingertip tactile sensors.
    Right: real-world manipulation tasks requiring diverse patterns of motion—contact coordination.
    \vspace{-4mm}
    }
    \label{fig:tasks}
\end{figure}

\section{Experiments}

Our experiments examine whether interaction latents provide a shared action space for learning, evaluating, and improving contact-rich manipulation.
We address four questions:
(1) Does ILA improve manipulation performance across policy architectures (Sec.~\ref{sec:main_results})?
(2) What motion and contact information do the learned representations preserve (Sec.~\ref{sec:representation})?
(3) Can the same latent space support interaction evaluation and offline policy improvement (Sec.~\ref{sec:offline_improvement_eval})?
(4) How do individual design choices contribute to these capabilities (Sec.~\ref{sec:ablations})?

\subsection{Experimental Setup}
\label{sec:setup}

\paragraph{Evaluation platform and benchmarks.}
We conduct all experiments on a dual-arm ARX X5 platform with parallel
grippers and fingertip Xense tactile sensors. An Intel RealSense D405
wrist camera and an external workspace camera provide visual observations.
Images, tactile deformation fields, and robot proprioception are synchronized
at 30\,Hz.
We evaluate four real-world contact-rich manipulation tasks
(Fig.~\ref{fig:tasks}): cucumber peeling, egg flipping, celery cutting,
and plate wiping, spanning sustained contact, contact transitions,
forceful tool use, and contact-maintaining surface interaction.
We collect 150 expert demonstrations per task for policy training.
Each method is evaluated over 20 trials per task; success rate (SR) is scored per cut for celery cutting and per trial for the other tasks.
We use spatially aggregated tactile deformation as a resultant-force proxy, with $F_{\mathrm{ext}}$ L1 measuring its mean absolute deviation from expert reference profiles during execution.
For open-loop prediction on held-out data, $F_{\mathrm{pred}}$ L1 and Tac.\ L1 measure mean absolute errors in predicted force proxies and three-dimensional tactile deformation fields, respectively, relative to ground truth.
Task procedures and evaluation metrics are detailed in
Appendices~\ref{task} and~\ref{app:evaluation_metrics}.

\paragraph{Comparison methods.}
We compare action formulations across pretrained VLA and WAM backbones and flow-matching (FM) policies trained from scratch, then evaluate offline improvement in the ILA space. 
For each VLA or FM backbone $P$, we compare three variants with tactile conditioning:
\textbf{P-tac} generates motor action sequences;
\textbf{P-tac$^{*}$} jointly generates motor action and tactile sequences, inspired by joint action--force prediction in KineDex~\citep{zhang2025kinedex} and ForceMimic~\citep{liu2025forcemimic};
and \textbf{P-ILA} generates interaction latents decoded into executable actions.
We use $\pi_{0.5}$~\citep{pi05_2025} as the VLA backbone.
For WAMs, we compare WAM-ILA with \textbf{FastWAM-tac}, a tactile-conditioned adaptation of FastWAM~\citep{Yuan2026FastWAMDW}, and \textbf{DreamTac}~\citep{lou2026dream}, which jointly generates actions and tactile signals.
Supplementary open-loop comparisons evaluate contact predictions decoded from policy-generated ILA latents against those of P-tac$^{*}$ and DreamTac.
P-tac and P-tac$^{*}$ omit S\&D tokens, which are
included in P-ILA. The three variants share tactile/proprioceptive
histories and execution horizons within each backbone.
For offline policy improvement, we compare our update with implicit Q-learning (IQL)~\citep{Kostrikov2021OfflineRL} and RECAP~\citep{Intelligence202506AV}.
All methods operate in the same frozen ILA action space and use the same fixed rollout corpus.
Within each policy family, methods share expert data and evaluation conditions.
Training and baseline details appear in Appendices~\ref{app:implementation_details} and~\ref{app:offline_baselines}.

\subsection{Policy Learning with Interaction Latent Actions}
\label{sec:main_results}

\begin{table}[t]
\vspace{-3.5mm}
\centering
\caption{\textbf{Performance on four contact-rich manipulation tasks.}
We report task success rate (SR) and execution force-proxy error ($F_{\mathrm{ext}}$ L1, mean $\pm$ standard deviation) relative to expert references.}
\label{tab:ila_main_results}
\setlength{\tabcolsep}{3pt}
\renewcommand{\arraystretch}{1.1}
\resizebox{0.98\linewidth}{!}{%
\begin{tabular}{llcccccccc}
\toprule
 & & \multicolumn{2}{c}{Cucumber Peeling} & \multicolumn{2}{c}{Egg Flipping} & \multicolumn{2}{c}{Celery Cutting} & \multicolumn{2}{c}{Plate Wiping} \\
\cmidrule(lr){3-4}\cmidrule(lr){5-6}\cmidrule(lr){7-8}\cmidrule(lr){9-10}
Family & Method & SR $\uparrow$ & $F_{\mathrm{ext}}$ L1 $\downarrow$ & SR $\uparrow$ & $F_{\mathrm{ext}}$ L1 $\downarrow$ & SR $\uparrow$ & $F_{\mathrm{ext}}$ L1 $\downarrow$ & SR $\uparrow$ & $F_{\mathrm{ext}}$ L1 $\downarrow$ \\
\midrule
\multicolumn{10}{l}{\emph{Interaction latent actions}} \\
FM & FM-tac & 50\% & $2.65 \pm 0.44$ & 85\% & $1.25 \pm 0.84$ & 16\% & $10.48 \pm 0.51 $ & 65\% & $5.51 \pm 0.31$ \\
 & FM-tac$^{*}$ & 15\% & $2.65 \pm 0.49$ & 65\% & $0.79 \pm 0.76$ & 14\% & $6.06 \pm 1.57$ & 80\% & $2.87 \pm 0.50$ \\
 & \textbf{FM-ILA} & 75\% & $2.04 \pm 0.37$ & 95\% & $0.56 \pm 0.64$ & 50\% & $7.52 \pm 0.78$ & 100\% & $3.44 \pm 0.35$ \\
\cmidrule(lr){1-10}
VLA ($\pi_{0.5}$) & VLA-tac & 15\% & $2.58 \pm 0.43$ & 25\% & $2.27 \pm 1.00$ & 6\% & $7.82 \pm 1.56 $ & 75\% & $2.69 \pm 0.53$ \\
 & VLA-tac$^{*}$ & 10\% & $2.82 \pm 0.47$ & 15\% &  $2.76 \pm 0.81$ & 10\% &  $ 9.24 \pm 2.47$ & 50\% &  $3.03 \pm 0.60$ \\
 & \textbf{VLA-ILA} & 50\% & $2.28 \pm 0.41$ & 85\% & $0.38 \pm 0.54$ & 37\% &$7.22 \pm 1.17 $ & 95\% & $2.51 \pm 0.38$ \\
\cmidrule(lr){1-10}
WAM & FastWAM-tac & 70\% & $2.01 \pm 0.53$ & 20\% & $1.95 \pm 0.92$ & 41\% & $7.84 \pm 0.91$ & 80\% & $1.54 \pm 0.35$ \\
 & DreamTac & 20\% & $2.63 \pm 0.68$ & 25\% & $2.13 \pm 0.83$ & 18\% & $8.54 \pm 2.23$ & 70\% & $2.92 \pm 0.65$ \\
 & \textbf{WAM-ILA} & 70\% & $1.44 \pm 0.47$ & 80\% & $0.14 \pm 0.50$ & 45\% & $8.94 \pm 0.63 $ & 100\% & $1.33 \pm 0.49$ \\
\midrule
\multicolumn{10}{l}{\emph{Offline policy improvement in the interaction latent space}} \\
FM & FM-ILA + IQL & 40\% & $2.42 \pm 1.12$ & 90\% & $1.39 \pm 0.73$ & 14\% & $9.64 \pm 1.75$ & 70\% & $2.50 \pm 0.59$ \\
 & FM-ILA + RECAP & 50\% & $1.66 \pm 0.95$ & 85\% & $1.55 \pm 0.62$ & 21\% & $9.25 \pm 0.81$ & 80\% & $1.92 \pm 0.93$ \\
 & \textbf{FM-ILA + ours} & 85\% & $0.51 \pm 0.50$ & 100\% & $0.49 \pm 0.55$ & 56\% & $5.55 \pm 0.38 $ & 100\% & $0.72 \pm 0.34$ \\
\bottomrule
\end{tabular}}
\vspace{-3mm}
\end{table}

We first ask whether interaction latents provide an effective policy action space across different model families. As shown in
Table~\ref{tab:ila_main_results}, ILA consistently improves task success across FM, VLA, and WAM policies. Relative to the tactile-conditioned direct-action counterparts, ILA improves average SR by 26.0, 36.5, and 21.0 percentage points for FM, VLA, and WAM, respectively, and achieves the highest or tied-highest SR within each family on all four tasks. 

ILA reduces $F_{\mathrm{ext}}$ L1 relative to the tactile-conditioned direct-action baselines in 11 of 12 comparisons.
On peeling, WAM-ILA matches FastWAM-tac's 70\% SR while reducing the error from 2.01 to 1.44.
For WAM on cutting, higher SR accompanies greater force-proxy error.
Thus, task completion and agreement with expert force-proxy profiles capture complementary aspects of performance.

\subsection{Motion—Contact Representation Analysis}
\label{sec:representation}
\begin{figure}[h]
    \vspace{-2mm}
    \centering
    \includegraphics[width=1.0\linewidth]{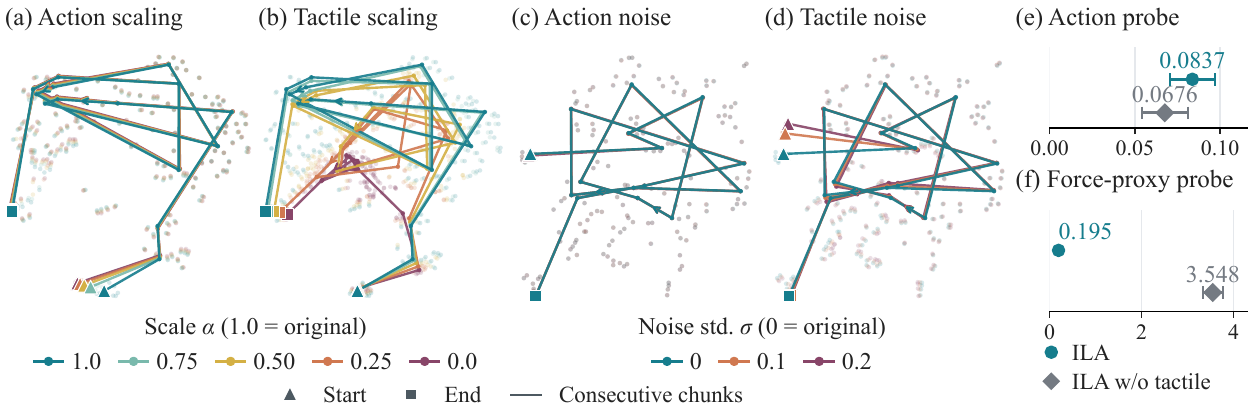}
    \caption{\textbf{Motion--contact representation analysis on cucumber peeling.}
    (a,b) Latent responses to action and tactile scaling;
    (c,d) responses to Gaussian input noise;
    (e,f) matched linear probes for action and force-proxy magnitude.
    Points denote latent states and lines trace one trajectory.
    Markers show mean L1 errors, with error bars indicating 95\% trajectory-bootstrap intervals.}
    \vspace{-4mm}
    \label{fig:ila_representation_analysis}
\end{figure}

We examine whether interaction latents capture both motion and contact
information using controlled interventions and linear probing
(Fig.~\ref{fig:ila_representation_analysis}).
On held-out cucumber-peeling trajectories, we independently scale or perturb
action and tactile inputs while keeping the remaining inputs fixed.
Scaling either modality changes the latent trajectories, with tactile scaling
inducing more pronounced shifts, whereas the trajectories remain largely
overlapping under Gaussian noise.
Linear probes assess motion and contact information retained in frozen
interaction-encoder outputs.
Compared with ILA w/o tactile, ILA reduces force-proxy magnitude L1 from
3.548 to 0.195, while action L1 increases from 0.0676 to 0.0837.
This contrast supports ILA's ability to retain contact-related information
beyond motion readout alone.
Results in Appendix~\ref{app:contact_prediction} further show that
policy-generated latents preserve predictive information about future tactile deformation and force proxies.

\subsection{Interaction Evaluation and Offline Policy Improvement}
\label{sec:offline_improvement_eval}

We evaluate offline policy improvement in the ILA space and examine the
reward and correction mechanisms that drive it.
With the ILA representation fixed, our offline improvement method achieves
higher SR and lower $F_{\mathrm{ext}}$ L1 than the IQL and RECAP adaptations
across all four tasks (Table~\ref{tab:ila_main_results}).
Notably, the improvement consistently reduces $F_{\mathrm{ext}}$ L1 across
all four tasks, indicating closer agreement with expert force-application
profiles.

\setlength{\intextsep}{4pt}
\setlength{\columnsep}{10pt}

\begin{wraptable}{r}{0.41\linewidth}
    \centering
    \small
    \setlength{\abovecaptionskip}{0pt}
    \setlength{\belowcaptionskip}{4pt}
    \caption{\textbf{Reward and correction ablations
    on cucumber peeling.}}
    \label{tab:ila_reward_ablation}

    \setlength{\tabcolsep}{2pt}
    \renewcommand{\arraystretch}{1.0}
    \begin{tabular*}{\linewidth}
        {@{\extracolsep{\fill}}lcc@{}}
        \toprule
        Setting & SR $\uparrow$
        & $F_{\mathrm{ext}}$ L1 $\downarrow$ \\
        \midrule
        \multicolumn{3}{@{}l}{\emph{Reward alternatives}} \\
        Grasp-intensity
            & 70\% & $1.78 \pm 0.48$ \\
        HapticVLA-style
            & 70\% & $1.69 \pm 0.58$ \\
        \midrule
        w/o failure correction
            & 80\% & $2.89 \pm 0.43$ \\
        \midrule
        \textbf{Ours}
            & \textbf{85\%}
            & $\mathbf{0.51} \pm 0.50$ \\
        \bottomrule
    \end{tabular*}
\end{wraptable}

\begin{figure}[h]
    \centering
    \includegraphics[width=1.0\linewidth]{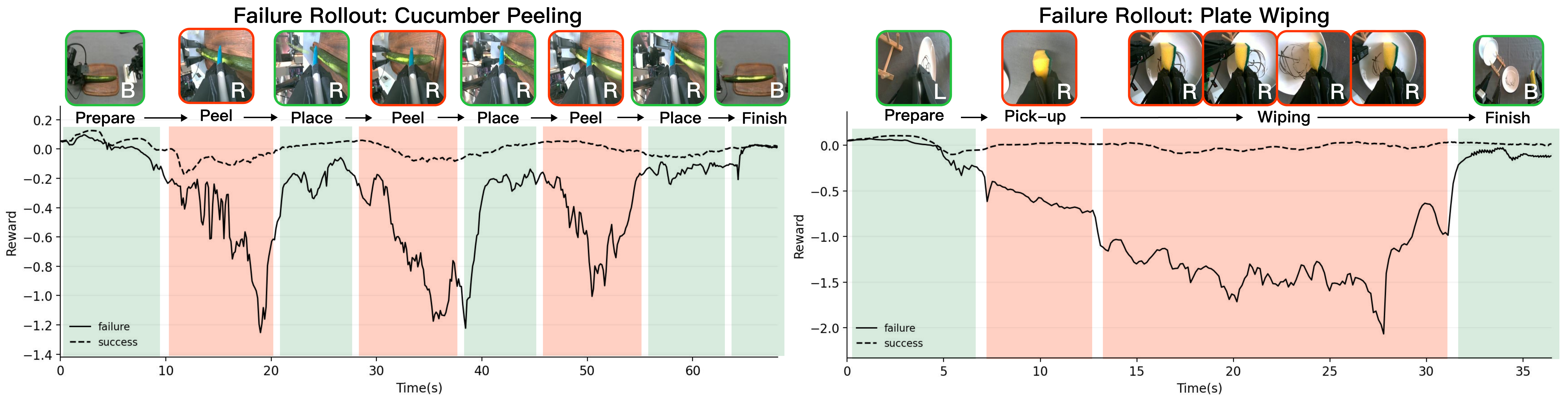}
    \vspace{-7mm}
    \caption{\textbf{Interaction rewards during manipulation.}
    Solid and dashed curves denote failed and successful rollouts,
    respectively. Reward decreases coincide with local interaction
    failures (red regions), while differences are smaller during
    phases with little interaction demand (green regions).}
    \label{fig:reward}
\end{figure}

\noindent
\textbf{Interaction reward assignment.}
We first visualize how the expert-consistency reward is assigned to individual
interaction chunks.
As shown in Fig.~\ref{fig:reward}, rewards drop during local interaction
failures in peeling and wiping, while successful rollouts maintain relatively
higher rewards; the gap narrows during phases with little interaction demand.
These examples illustrate temporally localized evaluation of interaction
consistency.

\smallskip
\noindent
\textbf{Reward and correction effectiveness.}
We further examine how reward design and failure correction contribute to
policy improvement (Table~\ref{tab:ila_reward_ablation}).
On peeling, replacing the expert-consistency reward with grasp-intensity
or HapticVLA-style tactile rewards~\citep{yu2026omnitactune,gubernatorov2026hapticvla}
reduces SR to 70\%.
Using our reward without failure retargeting reaches 80\% SR, while adding
reference-guided correction further improves SR to 85\% and reduces
$F_{\mathrm{ext}}$ L1 from 2.89 to 0.51.
Together, these results support expert-consistency evaluation and
reference-guided correction as complementary components of offline
improvement in the ILA space.

\begin{table}[h]
\centering
\caption{\textbf{Ablations of interaction representation, policy conditioning,
and action decoding.}
Results are reported on cucumber peeling and celery cutting.
All VLA variants retain native policy vision and tactile conditioning.}
\label{tab:ila_component_ablation}

\small
\setlength{\tabcolsep}{4pt}
\renewcommand{\arraystretch}{1.08}

\resizebox{0.76\linewidth}{!}{%
\begin{tabular}{@{}llcccc@{}}
\toprule
& &
\multicolumn{2}{c}{Cucumber Peeling} &
\multicolumn{2}{c}{Celery Cutting} \\
\cmidrule(lr){3-4}\cmidrule(lr){5-6}
Family & Setting
& SR $\uparrow$ & $F_{\mathrm{ext}}$ L1 $\downarrow$
& SR $\uparrow$ & $F_{\mathrm{ext}}$ L1 $\downarrow$ \\
\midrule

FM
& \textbf{Full ILA}
& \textbf{75\%} & $\mathbf{2.04} \pm 0.37$
& \textbf{50\%} & $\mathbf{7.52} \pm 0.78$ \\
& w/o implicit decoding
& 65\% & $2.32 \pm 0.53$
& 44\% & $7.97 \pm 2.45$ \\

\midrule

VLA ($\pi_{0.5}$)
& \textbf{Full ILA}
& \textbf{50\%} & $\mathbf{2.28} \pm 0.41$
& \textbf{37\%} & $7.22 \pm 1.17$ \\
& w/o S\&D
& 40\% & $3.13 \pm 0.58$
& 32\% & $\mathbf{6.51} \pm 0.92$ \\
& w/o S\&D + encoder vision
& 30\% & $2.78 \pm 0.46$
& 30\% & $8.68 \pm 1.11$ \\
& w/o tactile representation
& 20\% & $2.64 \pm 0.38$
& 10\% & $8.04 \pm 1.93$ \\

\bottomrule
\vspace{-4mm}
\end{tabular}%
}
\end{table}

\subsection{Ablation Studies}
\label{sec:ablations}

We ablate three components along the ILA pipeline: interaction
representation, policy conditioning, and action decoding
(Table~\ref{tab:ila_component_ablation}).
First, removing tactile inputs and supervision during latent learning reduces VLA success by 30 and 27 percentage points on peeling and cutting, respectively, and increases force-proxy error on both tasks.
Since policy tactile conditioning is retained, these results support incorporating contact into the learned action representation beyond using touch as an observation.
Second, we remove policy semantics-and-dynamics (S\&D) tokens either alone or together with encoder visual conditioning, while retaining native policy vision.
Both ablations reduce success on both tasks, supporting the additional visual context for interaction generation.
Removing policy S\&D tokens alone lowers SR by 10 and 5 percentage points on peeling and cutting, respectively, although cutting $F_{\mathrm{ext}}$ L1 improves from 7.22 to 6.51.
Finally, the full FM model improves SR over the decoder ablation by 10 and 6 percentage points on peeling and cutting, respectively, while reducing force-proxy error on both tasks.
These results support the implicit decoder design for learning executable interaction latents.

\section{Conclusion}
In this paper, we presented ILA, which uses decodable interaction latents as policy action units for contact-rich manipulation. Visually conditioned joint action—tactile reconstruction encodes motion, contact, and force-related deformation in a shared representation. These latents support VLA and WAM post-training and flow-matching policy learning from scratch. Tactile-conditioned expert-consistency rewards and observation-compatible reference correction further enable offline policy improvement in the same space. Across four real-world tasks and three policy families, ILA improves average success rates by 21.0--36.5 percentage points over tactile-conditioned direct-action baselines. On cucumber peeling, offline improvement reduces execution force-proxy error by 75\% while increasing success from 75\% to 85\%. These findings support incorporating contact evolution into action targets and using the representation for action generation, evaluation, and refinement.

\newpage
\bibliography{iclr2027_conference}

@inproceedings{Wirnshofer2020ControllingCM,
  title = {Controlling Contact-Rich Manipulation Under Partial Observability},
  author = {Florian Wirnshofer and Philipp Sebastian Schmitt and Georg von Wichert and Wolfram Burgard},
  booktitle = {Proceedings of Robotics: Science and Systems},
  year = {2020},
  doi = {10.15607/RSS.2020.XVI.023},
  url = {https://roboticsproceedings.org/rss16/p023.html}
}

@inproceedings{brown2019extrapolating,
  title = {Extrapolating Beyond Suboptimal Demonstrations via Inverse Reinforcement Learning from Observations},
  author = {Brown, Daniel and Goo, Wonjoon and Nagarajan, Prabhat and Niekum, Scott},
  booktitle = {Proceedings of the 36th International Conference on Machine Learning},
  year = {2019},
  volume = {97},
  pages = {783--792},
  series = {Proceedings of Machine Learning Research},
  publisher = {PMLR},
  url = {https://proceedings.mlr.press/v97/brown19a.html}
}

@inproceedings{biza2025robot,
  title = {On-robot reinforcement learning with goal-contrastive rewards},
  author = {Biza, Ondrej and Weng, Thomas and Sun, Lingfeng and Schmeckpeper, Karl and Kelestemur, Tarik and Ma, Yecheng Jason and Platt, Robert and van de Meent, Jan-Willem and Wong, Lawson LS},
  booktitle = {2025 IEEE International Conference on Robotics and Automation (ICRA)},
  year = {2025},
  pages = {4797--4805},
  organization = {IEEE}
}

@inproceedings{zakka2022xirl,
  title = {{XIRL}: Cross-embodiment inverse reinforcement learning},
  author = {Zakka, Kevin and Zeng, Andy and Florence, Pete and Tompson, Jonathan and Bohg, Jeannette and Dwibedi, Debidatta},
  booktitle = {Conference on Robot Learning},
  year = {2022},
  pages = {537--546},
  organization = {PMLR}
}

@article{ma2022vip,
  title = {{VIP}: Towards universal visual reward and representation via value-implicit pre-training},
  author = {Ma, Yecheng Jason and Sodhani, Shagun and Jayaraman, Dinesh and Bastani, Osbert and Kumar, Vikash and Zhang, Amy},
  journal = {arXiv preprint arXiv:2210.00030},
  year = {2022},
  url = {https://arxiv.org/abs/2210.00030}
}

@inproceedings{fei2026srpo,
  title = {{SRPO}: Self-referential policy optimization for vision-language-action models},
  author = {Fei, Senyu and Wang, Siyin and Ji, Li and Li, Ao and Zhang, Shiduo and Liu, Liming and Hou, Jinlong and Gong, Jingjing and Zhao, Xianzhong and Qiu, Xipeng},
  booktitle = {Proceedings of the IEEE/CVF Conference on Computer Vision and Pattern Recognition},
  year = {2026},
  pages = {6718--6728}
}

@article{huang2024diffusion,
  title = {Diffusion imitation from observation},
  author = {Huang, Bo-Ruei and Yang, Chun-Kai and Lai, Chun-Mao and Wu, Dai-Jie and Sun, Shao-Hua},
  journal = {Advances in Neural Information Processing Systems},
  year = {2024},
  volume = {37},
  pages = {137190--137217}
}

@article{zhang2026dream2reward,
  title = {{Dream2Reward}: Transition-Alignment Reward Models from Positive Demonstrations for Robotic Manipulation},
  author = {Zhang, Haoyu and Zeng, Zecui and Wang, Bin and Li, Lusong and Lin, Liang and Cheng, Long},
  journal = {arXiv preprint arXiv:2608.18787},
  year = {2026},
  url = {https://arxiv.org/abs/2608.18787}
}

@article{escontrela2023video,
  title = {Video prediction models as rewards for reinforcement learning},
  author = {Escontrela, Alejandro and Adeniji, Ademi and Yan, Wilson and Jain, Ajay and Peng, Xue Bin and Goldberg, Ken and Lee, Youngwoon and Hafner, Danijar and Abbeel, Pieter},
  journal = {Advances in Neural Information Processing Systems},
  year = {2023},
  volume = {36},
  pages = {68760--68783}
}

@article{higuera2024sparsh,
  title = {{Sparsh}: Self-supervised touch representations for vision-based tactile sensing},
  author = {Higuera, Carolina and Sharma, Akash and Bodduluri, Chaithanya Krishna and Fan, Taosha and Lancaster, Patrick and Kalakrishnan, Mrinal and Kaess, Michael and Boots, Byron and Lambeta, Mike and Wu, Tingfan and others},
  journal = {arXiv preprint arXiv:2410.24090},
  year = {2024},
  url = {https://arxiv.org/abs/2410.24090}
}

@article{zhao2024transferable,
  title = {Transferable tactile transformers for representation learning across diverse sensors and tasks},
  author = {Zhao, Jialiang and Ma, Yuxiang and Wang, Lirui and Adelson, Edward H},
  journal = {arXiv preprint arXiv:2406.13640},
  year = {2024},
  url = {https://arxiv.org/abs/2406.13640}
}

@inproceedings{yang2024binding,
  title = {Binding touch to everything: Learning unified multimodal tactile representations},
  author = {Yang, Fengyu and Feng, Chao and Chen, Ziyang and Park, Hyoungseob and Wang, Daniel and Dou, Yiming and Zeng, Ziyao and Chen, Xien and Gangopadhyay, Rit and Owens, Andrew and others},
  booktitle = {2024 IEEE/CVF Conference on Computer Vision and Pattern Recognition (CVPR)},
  year = {2024},
  pages = {26330--26343},
  organization = {IEEE}
}

@article{feng2026anytouch,
  title = {{AnyTouch} 2: General optical tactile representation learning for dynamic tactile perception},
  author = {Feng, Ruoxuan and Zhou, Yuxuan and Mei, Siyu and Zhou, Dongzhan and Wang, Pengwei and Cui, Shaowei and Fang, Bin and Yao, Guocai and Hu, Di},
  journal = {arXiv preprint arXiv:2602.09617},
  year = {2026},
  url = {https://arxiv.org/abs/2602.09617}
}

@article{zang2026tacforesight,
  title = {{TacForeSight}: Force-Guided Tactile World Model for Contact-Rich Manipulation},
  author = {Zang, Yujie and Zheng, Yuhang and Nie, Xian and Zheng, Yupeng and Tian, Shuai and Gu, Songen and Gao, Chen and Wang, Zining and Yan, Shuicheng and Ding, Wenchao},
  journal = {arXiv preprint arXiv:2606.11184},
  year = {2026},
  url = {https://arxiv.org/abs/2606.11184}
}

@article{huang2026taf,
  title = {{TaF-VLA}: Tactile-Force Alignment in Vision-Language-Action Models for Force-aware Manipulation},
  author = {Huang, Yuzhe and Lin, Pei and Li, Wanlin and Li, Daohan and Li, Jiajun and Jiang, Jiaming and Xiao, Chenxi and Jiao, Ziyuan},
  journal = {arXiv preprint arXiv:2601.20321},
  year = {2026},
  url = {https://arxiv.org/abs/2601.20321}
}

@article{xue2025reactive,
  title = {Reactive diffusion policy: Slow-fast visual-tactile policy learning for contact-rich manipulation},
  author = {Xue, Han and Ren, Jieji and Chen, Wendi and Zhang, Gu and Fang, Yuan and Gu, Guoying and Xu, Huazhe and Lu, Cewu},
  journal = {arXiv preprint arXiv:2503.02881},
  year = {2025},
  url = {https://arxiv.org/abs/2503.02881}
}

@article{zheng2026omnivta,
  title = {{OmniVTA}: Visuo-tactile world modeling for contact-rich robotic manipulation},
  author = {Zheng, Yuhang and Gu, Songen and Li, Weize and Zheng, Yupeng and Zang, Yujie and Tian, Shuai and Li, Xiang and Hao, Ce and Gao, Chen and Liu, Si and others},
  journal = {arXiv preprint arXiv:2603.19201},
  year = {2026},
  url = {https://arxiv.org/abs/2603.19201}
}

@article{Yuan2026FastWAMDW,
  title = {{Fast-WAM}: Do World Action Models Need Test-time Future Imagination?},
  author = {Tianyuan Yuan and Zibin Dong and Yicheng Liu and Hang Zhao},
  journal = {arXiv preprint arXiv:2603.16666},
  year = {2026},
  url = {https://arxiv.org/abs/2603.16666}
}

@article{lou2026dream,
  title = {{Dream-Tac}: A Unified Tactile World Action Model for Contact-Rich Robot Manipulation},
  author = {Lou, Yunfan and Ye, Yifan and Fu, Yankai and Cen, Jun and Chi, Xiaowei and Lyu, Yaoxu and Jia, Peidong and Han, Sirui and Lu, Zhihe and Zhang, Shanghang},
  journal = {arXiv preprint arXiv:2606.08737},
  year = {2026},
  url = {https://arxiv.org/abs/2606.08737}
}

@article{tian2026vt,
  title = {{VT-WAM}: Visual-Tactile World Action Model for Contact-Rich Manipulation},
  author = {Tian, Shuai and Zheng, Yupeng and Zheng, Yuhang and Gu, Songen and Zang, Yujie and Qin, Yuxing and Li, Weize and Li, Haoran and Ding, Wenchao and Zhao, Dongbin},
  journal = {arXiv preprint arXiv:2607.02503},
  year = {2026},
  url = {https://arxiv.org/abs/2607.02503}
}

@article{jin2026tacwam,
  title = {{TacWAM}: Anchor-Guided World Action Model with Mechanics-Aware Tactile Prediction},
  author = {Jin, Lei and Ma, Yiding and Zhang, Xin and Gao, Chen and Wu, Wei and Li, Yong},
  journal = {arXiv preprint arXiv:2607.28391},
  year = {2026},
  url = {https://arxiv.org/abs/2607.28391}
}

@article{Lee2024BehaviorGW,
  title = {Behavior Generation with Latent Actions},
  author = {Seungjae Lee and Yibin Wang and Haritheja Etukuru and H. J. Kim and Nur Muhammad Mahi Shafiullah and Lerrel Pinto},
  journal = {arXiv preprint arXiv:2403.03181},
  year = {2024},
  url = {https://arxiv.org/abs/2403.03181}
}

@article{Intelligence202506AV,
  title = {{$\pi_{0.6}^{*}$}: a {VLA} That Learns From Experience},
  author = {{Physical Intelligence} and Ali Amin and Raichelle Aniceto and Ashwin Balakrishna and Kevin Black and Ken Conley and Grace Connors and James Darpinian and Karan Dhabalia and Jared DiCarlo and Danny Driess and Michael Robert Equi and Adnan Esmail and Yun-Hao Fang and Chelsea Finn and Catherine Glossop and Thomas Godden and Ivan Goryachev and Lachy Groom and Hunter Hancock and Karol Hausman and Gashon Hussein and Brian Ichter and Szymon Jakubczak and Rowan Jen and Tim Jones and Ben Katz and Liyiming Ke and Chandra Kuchi and Marinda Lamb and Devin LeBlanc and Sergey Levine and Adrian Li-Bell and Yao Lu and Vishnu Mano and Mohith Mothukuri and Suraj Nair and Karl Pertsch and Allen Z. Ren and Charvi Sharma and Lucy Xiaoyang Shi and Laura Smith and Jost Tobias Springenberg and Kyle Stachowicz and Will Stoeckle and Alex Swerdlow and James Tanner and Marcel Torne and Quan Vuong and Anna Walling and Hao-Huan Wang and Blake Williams and Sukwon Yoo and Li-Li Yu and Ury Zhilinsky and Zhi-Yuan Zhou},
  journal = {arXiv preprint arXiv:2511.14759},
  year = {2025},
  url = {https://arxiv.org/abs/2511.14759}
}

@article{Kostrikov2021OfflineRL,
  title = {Offline Reinforcement Learning with Implicit Q-Learning},
  author = {Ilya Kostrikov and Ashvin Nair and Sergey Levine},
  journal = {arXiv preprint arXiv:2110.06169},
  year = {2021},
  url = {https://arxiv.org/abs/2110.06169}
}

@article{wan2026adversarial,
  title = {Adversarial Dual On-Policy Distillation from Expressive Teacher},
  author = {Wan, Zhenglin and Wu, Jingxuan and Yu, Xingrui and Zhang, Chubin and Lei, Mingcong and An, Bo and Tsang, Ivor W. and You, Yang},
  journal = {arXiv preprint arXiv:2605.27095},
  year = {2026},
  url = {https://arxiv.org/abs/2605.27095}
}

@article{Niu2026TRexTD,
  title = {{T-Rex}: Tactile-Reactive Dexterous Manipulation},
  author = {Dantong Niu and Zhuoyang Liu and Zekai Wang and Boning Shao and Zhao-Heng Yin and Anirudh Pai and Yuvan Sharma and Stefano Saravalle and Ruijie Zheng and Jing Wang and Ryan Punamiya and Mengda Xu and Yuqi Xie and Yunfan Jiang and Letian Fu and Konstantinos Kallidromitis and Matteo Di Gioia and Junyi Zhang and Jiaxin Ge and Haiwen Feng and Fabio Galasso and Wei Zhan and David M. Chan and Yutong Bai and Roei Herzig and Jiahui Lei and Fei-Fei Li and Ken Goldberg and Jitendra Malik and Pieter Abbeel and Yuke Zhu and Danfei Xu and Jim Fan and Trevor Darrell},
  journal = {arXiv preprint arXiv:2606.17055},
  year = {2026},
  url = {https://arxiv.org/abs/2606.17055}
}

@article{Feng2025AnyTouchLU,
  title = {{AnyTouch}: Learning Unified Static-Dynamic Representation across Multiple Visuo-tactile Sensors},
  author = {Ruoxuan Feng and Jiangyu Hu and Wenke Xia and Tian-Ci Gao and Ao Shen and Yuhao Sun and Bin Fang and Di Hu},
  journal = {arXiv preprint arXiv:2502.12191},
  year = {2025},
  url = {https://arxiv.org/abs/2502.12191}
}

@article{Tang2025ReLAMLA,
  title = {{ReLAM}: Learning Anticipation Model for Rewarding Visual Robotic Manipulation},
  author = {Nan Tang and Jing-Cheng Pang and Guanlin Li and Chao Qian and Yang Yu},
  journal = {arXiv preprint arXiv:2509.22402},
  year = {2025},
  url = {https://arxiv.org/abs/2509.22402}
}

@article{Nair2020AcceleratingOR,
  title = {Accelerating Online Reinforcement Learning with Offline Datasets},
  author = {Ashvin Nair and Murtaza Dalal and Abhishek Gupta and Sergey Levine},
  journal = {arXiv preprint arXiv:2006.09359},
  year = {2020},
  url = {https://arxiv.org/abs/2006.09359}
}

@article{Peng2019AdvantageWeightedRS,
  title = {Advantage-Weighted Regression: Simple and Scalable Off-Policy Reinforcement Learning},
  author = {Xue Bin Peng and Aviral Kumar and Grace Zhang and Sergey Levine},
  journal = {arXiv preprint arXiv:1910.00177},
  year = {2019},
  url = {https://arxiv.org/abs/1910.00177}
}

@article{Lipman2022FlowMF,
  title = {Flow Matching for Generative Modeling},
  author = {Yaron Lipman and Ricky T. Q. Chen and Heli Ben-Hamu and Maximilian Nickel and Matt Le},
  journal = {arXiv preprint arXiv:2210.02747},
  year = {2022},
  url = {https://arxiv.org/abs/2210.02747}
}

@inproceedings{Vaswani2017AttentionIA,
  title = {Attention is All you Need},
  author = {Ashish Vaswani and Noam Shazeer and Niki Parmar and Jakob Uszkoreit and Llion Jones and Aidan N. Gomez and Lukasz Kaiser and Illia Polosukhin},
  booktitle = {Neural Information Processing Systems},
  year = {2017},
  url = {https://api.semanticscholar.org/CorpusID:13756489}
}

@inproceedings{Chen2020LearningCI,
  title = {Learning Continuous Image Representation With Local Implicit Image Function},
  author = {Chen, Yinbo and Liu, Sifei and Wang, Xiaolong},
  booktitle = {Proceedings of the IEEE/CVF Conference on Computer Vision and Pattern Recognition (CVPR)},
  year = {2021},
  pages = {8628--8638},
  url = {https://openaccess.thecvf.com/content/CVPR2021/html/Chen_Learning_Continuous_Image_Representation_With_Local_Implicit_Image_Function_CVPR_2021_paper.html}
}

@inproceedings{Lee2018MakingSO,
  title = {Making Sense of Vision and Touch: Self-Supervised Learning of Multimodal Representations for Contact-Rich Tasks},
  author = {Lee, Michelle A. and Zhu, Yuke and Srinivasan, Krishnan and Shah, Parth and Savarese, Silvio and Fei-Fei, Li and Garg, Animesh and Bohg, Jeannette},
  booktitle = {2019 International Conference on Robotics and Automation (ICRA)},
  year = {2019},
  pages = {8943--8950},
  publisher = {IEEE},
  doi = {10.1109/ICRA.2019.8793485},
  url = {https://doi.org/10.1109/ICRA.2019.8793485}
}

@article{Dong2026ForceDeltaVLADF,
  title = {{ForceDelta-VLA}: Distilling Force-Conditioned ActionCorrections for Contact-Rich Manipulation},
  author = {Ju Dong and Yu Fu and Jian Chen and Yimeng Liu and Haocheng Zhao and Lei Zhang and Kaixin Bai and Liding Zhang and Diwen Zheng and Alois Christian Knoll and Angela P. Schoellig and Jianwei Zhang},
  journal = {arXiv preprint arXiv:2609.18242},
  year = {2026},
  url = {https://arxiv.org/abs/2609.18242}
}

@article{Oquab2023DINOv2LR,
  title = {{DINOv2}: Learning Robust Visual Features without Supervision},
  author = {Maxime Oquab and Timoth{\'e}e Darcet and Th{\'e}o Moutakanni and Huy V. Vo and Marc Szafraniec and Vasil Khalidov and Pierre Fernandez and Daniel Haziza and Francisco Massa and Alaaeldin El-Nouby and Mido Assran and Nicolas Ballas and Wojciech Galuba and Russ Howes and Po-Yao (Bernie) Huang and Shang-Wen Li and Ishan Misra and Michael G. Rabbat and Vasu Sharma and Gabriel Synnaeve and Hu Xu and Herv{\'e} J{\'e}gou and Julien Mairal and Patrick Labatut and Armand Joulin and Piotr Bojanowski},
  journal = {arXiv preprint arXiv:2304.07193},
  year = {2023},
  url = {https://arxiv.org/abs/2304.07193}
}

@article{Qin2026DexTouchWMLA,
  title = {{DexTouch-WM}: Learning Action-Conditioned Tactile World Models from Human Touch for Dexterous Robot Manipulation},
  author = {Yan Qin and Yue Chen and Wenwei Lin and Shujia Liu and Chuqiao Lyu and Kailun Su and Weiyang Jin and Chenze Yu and Ping Luo and Wenbo Ding and Tianxing Chen and Renjing Xu},
  journal = {arXiv preprint arXiv:2609.20649},
  year = {2026},
  url = {https://arxiv.org/abs/2609.20649}
}

@article{Zhang2026ContactWorldWR,
  title = {{ContactWorld}: What Representations Matter for Vision-Tactile Latent World Models in Contact-Rich Manipulation},
  author = {Zhiyuan Zhang and Pokuang Zhou and Kaidi Zhang and Adeesh Desai and Temitope Amosa and Davood Soleymanzadeh and Jiuzhou Lei and Yuhao Zhou and Minghui Zheng and Yu She},
  journal = {arXiv preprint arXiv:2606.13877},
  year = {2026},
  url = {https://arxiv.org/abs/2606.13877}
}

@article{Zhang2026ImaginingTS,
  title = {Imagining the Sense of Touch: Touch-Informed Manipulation via Imagined Tactile Representations},
  author = {Zhi-Yuan Zhang and Adeesh Desai and Jyun-Chi Hu and Yosuke Saka and Quan Khanh Luu and Jiuzhou Lei and Davood Soleymanzadeh and Bi-Hao Zhang and Ming-Hui Zheng and Yu She},
  journal = {arXiv preprint arXiv:2607.01684},
  year = {2026},
  url = {https://arxiv.org/abs/2607.01684}
}

@article{Zhou2026TouchWorldAP,
  title = {{TouchWorld}: A Predictive and Reactive Tactile Foundation Model for Dexterous Manipulation},
  author = {Jianyi Zhou and Feiyang Hong and Yun-Hao Li and Yicheng Zhao and Yongjue Cen and Zirui Liu and Jia-Kang Huang and Zirui Chen and Ruiyang Zhang and Weizhuo Zhu and Xuhua Song and Shuo Yang},
  journal = {arXiv preprint arXiv:2607.07287},
  year = {2026},
  url = {https://arxiv.org/abs/2607.07287}
}

@article{Xu2025exUMIER,
  title = {{exUMI}: Extensible Robot Teaching System with Action-aware Task-agnostic Tactile Representation},
  author = {Yue Xu and Litao Wei and Pengyu An and Qingyu Zhang and Yong-Lu Li},
  journal = {arXiv preprint arXiv:2509.14688},
  year = {2025},
  url = {https://arxiv.org/abs/2509.14688}
}

@article{team2026n_0,
  title = {{$N_0$-TWAM}: Scaling Tactile-Native World-Action Model for Contact-Rich Manipulation},
  author = {{NeoteAI Team} and {Fudan TEAI Team}},
  journal = {arXiv preprint arXiv:2607.23783},
  year = {2026},
  url = {https://arxiv.org/abs/2607.23783}
}

@article{zhang2025kinedex,
  title = {{KineDex}: Learning tactile-informed visuomotor policies via kinesthetic teaching for dexterous manipulation},
  author = {Zhang, Di and Yuan, Chengbo and Wen, Chuan and Zhang, Hai and Zhao, Junqiao and Gao, Yang},
  journal = {arXiv preprint arXiv:2505.01974},
  year = {2025},
  url = {https://arxiv.org/abs/2505.01974}
}

@inproceedings{liu2025forcemimic,
  title = {{ForceMimic}: Force-Centric Imitation Learning with Force-Motion Capture System for Contact-Rich Manipulation},
  author = {Liu, Wenhai and Wang, Junbo and Wang, Yiming and Wang, Weiming and Lu, Cewu},
  booktitle = {2025 IEEE International Conference on Robotics and Automation (ICRA)},
  year = {2025},
  pages = {1105--1112},
  publisher = {IEEE},
  doi = {10.1109/ICRA55743.2025.11128061},
  url = {https://forcemimic.github.io/}
}

@article{zhang2026me,
  title = {{ME-Dex} 1.0: Bringing Heterogeneous Tactile Sensing into World Action Modeling},
  author = {Zhang, Xuancheng and Liu, Xuetao and Tang, Qianying and Wang, Jizhe and Wang, Runsheng and Cheng, Zhijing and Lin, Bochen and Wen, Haoran and Li, Ming and Zhan, Kun and others},
  journal = {arXiv preprint arXiv:2609.21449},
  year = {2026},
  url = {https://arxiv.org/abs/2609.21449}
}

@inproceedings{chen2025sarm,
  title = {{SARM}: Stage-Aware Reward Modeling for Long Horizon Robot Manipulation},
  author = {Chen, Qianzhong and Yu, Justin and Schwager, Mac and Abbeel, Pieter and Shentu, Yide and Wu, Philipp},
  booktitle = {International Conference on Learning Representations},
  year = {2026},
  url = {https://arxiv.org/abs/2509.25358}
}

@inproceedings{Huang2023DiffusionReward,
  title = {Diffusion Reward: Learning Rewards via Conditional Video Diffusion},
  author = {Huang, Tao and Jiang, Guangqi and Ze, Yanjie and Xu, Huazhe},
  booktitle = {European Conference on Computer Vision},
  year = {2024},
  url = {https://arxiv.org/abs/2312.14134}
}

@inproceedings{lai2024drail,
  title = {Diffusion-Reward Adversarial Imitation Learning},
  author = {Lai, Chun-Mao and Wang, Hsiang-Chun and Hsieh, Ping-Chun and Wang, Yu-Chiang Frank and Chen, Min-Hung and Sun, Shao-Hua},
  booktitle = {Advances in Neural Information Processing Systems},
  year = {2024},
  volume = {37},
  pages = {95456--95487},
  publisher = {Curran Associates, Inc.},
  doi = {10.52202/079017-3025},
  url = {https://proceedings.neurips.cc/paper_files/paper/2024/hash/ad47b1801557e4be37d30baf623de426-Abstract-Conference.html}
}

@article{gubernatorov2026hapticvla,
  title = {{HapticVLA}: Contact-Rich Manipulation via Vision-Language-Action Model without Inference-Time Tactile Sensing},
  author = {Konstantin Gubernatorov and Mikhail Sannikov and Ilya Mikhalchuk and Egor Kuznetsov and Makar Artemov and Ogunwoye Faith Oluwatobi and Marcelino Fernando and Artem Asanov and Ziang Guo and Dzmitry Tsetserukou},
  journal = {arXiv preprint arXiv:2603.15257},
  year = {2026},
  url = {https://arxiv.org/abs/2603.15257}
}

@article{yu2026omnitactune,
  title = {{OmniTacTune}: Policy-Agnostic Real-World {RL} for Tactile Residual Adaptation of Visual Policies},
  author = {Kelin Yu and Haode Zhang and Harish Ravichandar and Yunhai Han and Ruohan Gao},
  journal = {arXiv preprint arXiv:2607.03723},
  year = {2026},
  url = {https://arxiv.org/abs/2607.03723}
}

@article{pi05_2025,
  title = {{$\pi_{0.5}$}: a Vision-Language-Action Model with Open-World Generalization},
  author = {{Physical Intelligence} and Kevin Black and Noah Brown and James Darpinian and Karan Dhabalia and Danny Driess and Adnan Esmail and Michael Equi and Chelsea Finn and Niccolo Fusai and Manuel Y. Galliker and Dibya Ghosh and Lachy Groom and Karol Hausman and Brian Ichter and Szymon Jakubczak and Tim Jones and Liyiming Ke and Devin LeBlanc and Sergey Levine and Adrian Li-Bell and Mohith Mothukuri and Suraj Nair and Karl Pertsch and Allen Z. Ren and Lucy Xiaoyang Shi and Laura Smith and Jost Tobias Springenberg and Kyle Stachowicz and James Tanner and Quan Vuong and Homer Walke and Anna Walling and Haohuan Wang and Lili Yu and Ury Zhilinsky},
  journal = {arXiv preprint arXiv:2504.16054},
  year = {2025},
  url = {https://arxiv.org/abs/2504.16054}
}

@article{bi2026vla,
  title={Vla-touch: Enhancing vision-language-action model with dual-level tactile feedback},
  author={Bi, Jianxin and Ma, Kevin Yuchen and Hao, Ce and Zheng, Mike Shou and Soh, Harold},
  journal={IEEE Robotics and Automation Letters},
  year={2026},
  publisher={IEEE}
}

@article{he2025foar,
  title={Foar: Force-aware reactive policy for contact-rich robotic manipulation},
  author={He, Zihao and Fang, Hongjie and Chen, Jingjing and Fang, Hao-Shu and Lu, Cewu},
  journal={IEEE Robotics and Automation Letters},
  volume={10},
  number={6},
  pages={5625--5632},
  year={2025},
  publisher={IEEE}
}

@article{huang20243d,
  title={3d-vitac: Learning fine-grained manipulation with visuo-tactile sensing},
  author={Huang, Binghao and Wang, Yixuan and Yang, Xinyi and Luo, Yiyue and Li, Yunzhu},
  journal={arXiv preprint arXiv:2410.24091},
  year={2024}
}
\bibliographystyle{iclr2027_conference}

\newpage
\appendix
\clearpage
\section*{Appendix}
\pdfbookmark[0]{Appendix}{ila.appendix.contents}

\begingroup
\normalsize
\setlength{\parindent}{0pt}
\setlength{\parskip}{0pt}
\makeatletter
\c@tocdepth=2\relax
\def\@dotsep{3}
\newcommand{\ILAappsection}[2]{%
  \vspace{8pt}%
  \l@section
    {\hyperref[#1]{\numberline{\ref*{#1}}#2}}%
    {\hyperref[#1]{\pageref*{#1}}}%
  \vspace{4pt}%
}
\newcommand{\ILAappsubsection}[2]{%
  \l@subsection
    {\hyperref[#1]{\numberline{\ref*{#1}}#2}}%
    {\hyperref[#1]{\pageref*{#1}}}%
  \vspace{8pt}%
}
\makeatother

\ILAappsection{app:appendix}{Additional Method and Training Details}
\ILAappsubsection{app:implementation_details}{Representation Training and Policy Integration}
\ILAappsubsection{app:improvement_interpretation}{Interpretation of Expert-Reference-Guided Improvement}
\ILAappsubsection{app:successful_rollout_reweighting}{Successful Rollout Reweighting}
\ILAappsubsection{app:reference_selection}{Observation-Compatible Reference Selection}
\ILAappsubsection{app:offline_update_implementation}{Offline Update Objective and Optimization}

\ILAappsection{app:experimental_details}{Experimental Details and Additional Results}
\ILAappsubsection{task}{Data Collection and Task Procedures}
\ILAappsubsection{app:evaluation_metrics}{Evaluation Metrics}
\ILAappsubsection{app:offline_baselines}{Offline Policy-Improvement Baselines}
\ILAappsubsection{app:reward_design}{Reward-Design Comparisons}
\ILAappsubsection{app:representation_perturbations}{Additional Motion--Contact Representation Analysis}
\ILAappsubsection{app:representation_visualization}{Qualitative Visualization of Interaction Representations}
\ILAappsubsection{app:contact_prediction}{Open-Loop Tactile and Force-Proxy Prediction}
\ILAappsubsection{app:trajectory_replay}{Comparison with Action-Trajectory Replay}
\ILAappsubsection{app:offline_diagnostics}{Diagnostics of Offline Policy Improvement}

\ILAappsection{app:shared_interaction_latents}{Towards a Shared Interaction Latent Space}

\ILAappsection{app:rollout_cases}{Rollout Cases and Interaction Analysis}

\ILAappsection{app:ai_use_statement}{AI Use Statement}

\endgroup
\clearpage

\section{Additional Method and Training Details}
\label{app:appendix}

\subsection{Representation Training and Policy Integration}
\label{app:implementation_details}

We detail the representation architecture, training, and policy interfaces
introduced in Secs.~\ref{sec:interaction_representation} and~\ref{sec:interaction_policy}.
Offline baseline and reward-comparison settings are provided in
Appendices~\ref{app:offline_baselines} and~\ref{app:reward_design}, respectively.

\paragraph{Interaction representation pretraining.}
Action and tactile sequences are synchronized at 30\,Hz. Each action
specifies 14 absolute joint and gripper targets for the two arms, and each
of the four fingertip sensors provides a $35\times20$ deformation field
with three channels. We extract training chunks of
$H\in\{32,64,128\}$ frames with a one-frame sliding-window stride.
The encoder compresses time by a factor of four, producing
$K=H/4$ tokens with $d_z=128$. Policy learning uses $H=128$ and $K=32$.
The action encoder produces one token per frame, while the shared tactile
encoder produces two spatial tokens per sensor, yielding nine tokens per
frame before fusion. Learned temporal and modality-slot embeddings
preserve their identities. The encoder attends bidirectionally within
the target chunk and is visually conditioned on images available at the
chunk start. With $H=128$ and 256 DINOv2 patches per view, the representation
module contains approximately 10.03M parameters: 9.48M in the encoder,
including visual feature processing, and 0.55M in the reconstruction heads.
These counts exclude the frozen DINOv2 backbone and the policy network.

\paragraph{Continuous reconstruction and supervision.}
The decoder interpolates neighboring latent anchors, spaced four frames
apart, and concatenates the resulting feature with a learned time encoding.
It combines elapsed time, normalized progress, chunk duration, and
Fourier features at $\{0.5,1,2,4,8\}$\,Hz.
A shared MLP maps the concatenated 192-dimensional input through a
256-dimensional hidden layer to a 128-dimensional query feature, from
which separate heads reconstruct actions, tactile fields, and aggregate
deformation. The decoder requires no additional visual input, and its
tactile head does not depend on reconstructed actions.
This design separates policy inference from lightweight action decoding:
once a latent chunk has been generated, it can be reused to evaluate
actions at successive control timestamps without rerunning the visual
encoder or the generative policy. The deployed control rate also depends
on inference latency and the robot controller.

We supervise the $H$ observed times and $2H$ stratified inter-frame queries,
with interpolation weight $\lambda_{\mathrm{int}}=0.5$.
Actions and deformation fields use training-set normalization.
Aggregate deformation is computed in the original sensor coordinates
and scaled by a fixed per-sensor RMS statistic estimated from training
data. Targets at inter-frame queries are obtained by interpolating
adjacent measurements; they impose temporal consistency without adding
new high-frequency observations.

\paragraph{Decoder ablation.}
For the \emph{w/o implicit decoding} variant in
Table~\ref{tab:ila_component_ablation}, we replace the time-conditioned
implicit decoder with a discrete decoder comprising transposed
convolutions and MLPs, preserving the decoder parameter count.
We remove query-time encoding and inter-frame interpolation supervision,
and train reconstruction only at the original observed frames.
This ablation evaluates the combined contribution of continuous-time
decoding and inter-frame supervision.

\paragraph{Representation training schedule.}
We train the interaction representation for 70 epochs with a batch size
of 256. The learning rate follows a cosine decay from $2\times10^{-4}$
to $5\times10^{-6}$, reaching its minimum at epoch 50.

\paragraph{Policy integration.}
We use $\pi_{0.5}$~\citep{pi05_2025} and
FastWAM~\citep{Yuan2026FastWAMDW} as the VLA and WAM backbones,
respectively, and fine-tune all policy parameters.
The pretrained interaction encoder and decoder remain frozen during
policy learning. Policies predict standardized interaction latents,
which are denormalized before action decoding. For VLA adaptations,
we retain native image and language inputs and augment the VLM prefix
with visual dynamics, tactile, and proprioceptive tokens. For WAM
adaptations, these tokens augment the shared cross-attention context,
while the native video-generation objective is retained. In both cases,
the action branch generates interaction latents instead of raw action
sequences. Policy conditions use only observations available at the
chunk start.

\paragraph{Policy variant configurations.}
Table~\ref{tab:ila_variant_configs} summarizes the VLA and FM variants.
Within each backbone, all three variants use the same tactile and
proprioceptive histories, execute all 128 predicted actions before
replanning, and use the same sampling budget.
P-ILA additionally uses S\&D tokens from the frozen visual branch of
the interaction representation.
These comparisons evaluate the complete ILA policy configuration;
the contribution of S\&D conditioning is examined separately in
Sec.~\ref{sec:ablations}.

\begin{table}[t]
\centering
\small
\caption{\textbf{Configurations of the VLA and FM policy variants.}}
\label{tab:ila_variant_configs}
\setlength{\tabcolsep}{6pt}
\begin{tabular}{lll}
\toprule
Variant & S\&D tokens & Generation target \\
\midrule
P-tac & No & Motor actions \\
P-tac$^{*}$ & No & Motor actions and tactile fields \\
P-ILA & Yes & Interaction latents \\
\bottomrule
\end{tabular}
\end{table}

\paragraph{Policy training and execution.}
For Stage~2, all three policy families use AdamW, following
$\pi_{0.5}$~\citep{pi05_2025}, with a batch size of 128 and a learning rate
of $5\times10^{-5}$. We fine-tune WAM and VLA policies for 40,000 and
50,000 steps, respectively, and train FM policies from scratch for
100,000 steps. At each planning step, the policy generates $K=32$ latent
tokens, which the frozen decoder maps to $H=128$ motor actions on a
30\,Hz time grid. We execute all 128 actions before synchronously
replanning from updated observations.

\subsection{Interpretation of Expert-Reference-Guided Improvement}
\label{app:improvement_interpretation}

We specify the reward implementation and its interpretation for
the offline improvement in Sec.~\ref{sec:interaction_improvement}.

\paragraph{Nature of the expert reference.}
The reference bank contains only windows from expert demonstrations.
For each expert window, the frozen interaction encoder maps the recorded
action and tactile sequences, conditioned on their visual context, to
a reference latent
\[
z_j^e = \operatorname{Norm}
\bigl(E_\phi(A_j^e,S_j^e;V_j^e)\bigr).
\]
These encoded expert latents populate the reference bank and provide the
clean targets for reward-model training.
The policy does not generate or revise these reference targets. This construction grounds the reference distribution directly in recorded
expert interactions and makes it independent of policy prediction accuracy.

\paragraph{Rollout records and temporal alignment.}
Each rollout record $(o_i,\hat z_i,A_i,S_i)$ corresponds to one complete
execution chunk. Here $o_i$ contains the observations available at the
planning step, and $\hat z_i$ is the normalized 32-token policy prediction
generated from those observations. The sequence $A_i$ contains the
128 decoded motor commands issued to the robot, and $S_i$ contains the
tactile measurements recorded over the same 128 control steps.
The next policy prediction is generated only after this chunk has been
executed, so the stored prediction and measured contact share the same
execution interval.

\paragraph{Role of tactile conditioning.}
Each 128-frame interaction window is compressed into $K=32$ latent tokens,
with token $k$ corresponding to a four-frame interval.
For each interval, we compute a four-dimensional tactile descriptor
$c_{ik}\in\mathbb{R}^{4}$ from the tactile measurements recorded during
execution. We denote the aligned descriptor sequence by
$c_i=\{c_{ik}\}_{k=1}^{K}$.
The reward model conditions each latent token $\hat z_{ik}$ on its
corresponding descriptor $c_{ik}$, thereby evaluating the predicted
interaction relative to expert references under the observed contact
condition.
Importantly, these descriptors are used only for offline reward evaluation;
policy generation remains conditioned solely on observations available at
the beginning of the interaction window.

\paragraph{Reward network and score estimation.}
We use a FiLM-conditioned residual MLP (\texttt{film\_residual}) with
128-dimensional latent inputs, hidden width 768, and three residual
blocks. The four-dimensional tactile condition modulates the hidden
features through FiLM. The network evaluates each token independently.
For a candidate token $\hat z_{ik}$, define
\[
\ell_{ik}(s,\epsilon)=\frac{1}{d_z}
\left\|v_\omega((1-s)\epsilon+s\hat z_{ik},s,c_{ik})
-(\hat z_{ik}-\epsilon)\right\|_2^2.
\]
We evaluate the five flow times
$\mathcal S=\{0.1,0.25,0.5,0.75,1.0\}$ with paired Gaussian noises
$\epsilon_{ks}$ and $-\epsilon_{ks}$, using fixed seed 42.
The raw window reward averages ten residual evaluations per token:
\begin{equation}
r_i^{\mathrm{raw}}
=-\frac{1}{10K}\sum_{k=1}^{K}\sum_{s\in\mathcal S}
\left[\ell_{ik}(s,\epsilon_{ks})+\ell_{ik}(s,-\epsilon_{ks})\right].
\label{eq:implemented_window_reward}
\end{equation}
We standardize these rewards using fixed statistics from the successful
training windows before weighting and reference selection.
The same scoring rule evaluates encoded references using their aligned
recorded tactile conditions.
The divisors of 10 or 20 used to display reward curves do not enter
this training-reward computation.

\paragraph{Interpretation of the flow-matching residual.}
Let $Z \sim q_{\mathrm{ref}}(\cdot \mid c)$ denote a reference
latent token conditioned on its aligned tactile descriptor, and let
$\epsilon$ be independent standard Gaussian
noise. For the interpolation
$X_s = (1-s)\epsilon + sZ$, the population minimizer of the
squared flow-matching objective is the conditional mean
velocity~\citep{Lipman2022FlowMF}:
\begin{equation}
    v^\star(x,s,c)
    =
    \mathbb{E}\!\left[
        Z-\epsilon \mid X_s=x,c
    \right].
    \label{eq:reference_conditional_velocity}
\end{equation}
This is an average over compatible reference--noise pairs;
individual pairs can have nonzero residuals even at the
population optimum.

For $s<1$, define the implied clean-latent estimate
\begin{equation}
    m_\omega(x,s,c)
    :=
    x + (1-s)v_\omega(x,s,c).
    \label{eq:reference_latent_estimate}
\end{equation}
At the population optimum,
$m^\star(x,s,c)=\mathbb{E}[Z \mid X_s=x,c]$.
For a candidate token $\hat z_{ik}$, write
$x_s=(1-s)\epsilon+s\hat z_{ik}$ and let $D=d_z=128$.
Since $\hat z_{ik}=x_s+(1-s)(\hat z_{ik}-\epsilon)$, we obtain
\begin{equation}
    \frac{1}{D}
    \left\|
        v_\omega(x_s,s,c_{ik})-(\hat z_{ik}-\epsilon)
    \right\|_2^2
    =
    \frac{
        \left\|m_\omega(x_s,s,c_{ik})-\hat z_{ik}\right\|_2^2
    }{D(1-s)^2}.
    \label{eq:reference_reconstruction_interpretation}
\end{equation}
Thus, at these interpolation times, the velocity residual is
equivalent to a time-weighted latent reconstruction error.
It measures how consistently the learned reference field
recovers a candidate from its perturbed versions.
Here $m_\omega$ is an algebraic reparameterization of the
velocity field, not an additional learned decoder.
The evaluation distribution $\nu$ assigns equal mass to the five
times in $\mathcal S$.
At $s=1$, the original velocity residual remains well defined;
the reconstruction identity above does not apply.

\subsection{Successful Rollout Reweighting}
\label{app:successful_rollout_reweighting}

We specify the successful-rollout reweighting used in
Eq.~\eqref{eq:offline_policy_improvement}.
Although all records in $\mathcal D^+$ come from successful rollouts,
their expert-consistency rewards can vary within a trajectory.
The weighting rule uses this variation to adjust the contributions
of individual interaction chunks to policy learning.

For a successful interaction chunk with estimated reward
$\widehat r_i$, we use
\begin{equation}
w_i
=\eta+(1-\eta)
\frac{
\exp\!\left[
\operatorname{clip}\!\left(
\frac{\widehat r_i-m_+}{\beta\sigma_+},
-a_{\mathrm{clip}},a_{\mathrm{clip}}
\right)\right]
}{
\mathbb E_{j\sim P_+}\!\left[
\exp\!\left[
\operatorname{clip}\!\left(
\frac{\widehat r_j-m_+}{\beta\sigma_+},
-a_{\mathrm{clip}},a_{\mathrm{clip}}
\right)\right]
\right]
}.
\label{eq:success_weights}
\end{equation}
Here $P_+$ is the uniform empirical distribution over successful
training windows, and $m_+$ and $\sigma_+>0$ are the reward median
and standard-deviation scale of this pool.
These statistics and the denominator are fixed before improvement,
rather than recomputed within each minibatch.
We use $\eta=0.3$, $\beta=2$, and $a_{\mathrm{clip}}=3$ for all tasks.

The successful-rollout objective is the first term of
Eq.~\eqref{eq:offline_policy_improvement}:
\begin{equation}
\mathcal L_+(\theta)
=\mathbb E_{i\sim P_+}
\left[
w_i\,\ell_{\mathrm{FM}}(\theta;o_i,\hat z_i)
\right].
\label{eq:success_objective}
\end{equation}
Weights are nondecreasing in reward, and the uniform component gives
$w_i\geq0.3$.
Thus, each successful record retains a positive loss coefficient,
while clipping bounds the exponential component's response to
extreme reward values.

\subsection{Observation-Compatible Reference Selection}
\label{app:reference_selection}

We specify the construction of the reference pairs $\mathcal P$
used in Eq.~\eqref{eq:reference_correction}.
For a low-reward record from a failed rollout, we retrieve a
high-reward reference with a compatible observation and a direction
consistent with other retrieved candidates.
Each reference $z_j^e$ is the normalized interaction-encoder output of an
expert demonstration window, as defined in
Appendix~\ref{app:improvement_interpretation}.

\paragraph{Reward and observation compatibility.}
We consider failed-rollout records satisfying
$\widehat r_i<\tau_-$ and references satisfying
$\widehat r_j\geq\tau_+$.
Visual, proprioceptive, and tactile observation features are separately
projected to 64 dimensions and normalized, then concatenated into a
192-dimensional retrieval embedding $\psi(o)$.
We define
\begin{equation}
\delta_{ij}=\|\psi(o_i)-\psi(o_j)\|_2,
\qquad
d_{ij}=\delta_{ij}^2.
\label{eq:reference_observation_distance}
\end{equation}
Candidate retrieval uses $\delta_{ij}$, whereas the compatibility
coefficient in Eq.~\eqref{eq:reference_correction} uses $d_{ij}$.
Among reward-qualified references from the same task, data partition,
and observation-view configuration, we retain the nearest window
from each expert demonstration trajectory.
Temporally overlapping windows from one episode therefore contribute
at most one candidate to the subsequent consistency check.
There is no additional reward-gap or latent-distance threshold.
For egg flipping, failed source windows are restricted to start times
between 0 and 12.5\,s.

\paragraph{Directional consistency.}
For each retained candidate with $z_j^{\mathrm e}\neq\hat z_i$, define
\begin{equation}
\mathbf u_{ij}
=\frac{\operatorname{vec}(z_j^{\mathrm e}-\hat z_i)}
{\|z_j^{\mathrm e}-\hat z_i\|_F}.
\label{eq:reference_selection_direction}
\end{equation}
All differences use the same fixed normalized policy coordinates;
candidates with zero-length differences are excluded.
We connect candidates $j$ and $k$ when
$\mathbf u_{ij}^{\mathsf T}\mathbf u_{ik}\geq0.6$
and retain a largest group satisfying this all-pairs condition.
The threshold applies to every pair within the retained group.
Ties between groups of the same size are resolved by higher mean
pairwise cosine and then smaller mean observation distance.

\paragraph{Reference assignment.}
Sources are processed in decreasing group size, followed by
decreasing mean pairwise cosine and increasing mean observation
distance.
Within each group, references are ranked by their mean cosine to
the other members, with observation distance breaking ties.
The first reference with remaining capacity is assigned to the source.
Each reference window can be assigned to at most 18 sources within
a data partition; sources are omitted if every group member has
reached this limit.
Remaining ties are resolved using fixed record identifiers.
The capacity constraint is applied after group construction.

The resulting pair set $\mathcal P$ is fixed before policy improvement.
Directional consistency uses the stored prediction $\hat z_i$ as
its origin, whereas the correction target in
Eq.~\eqref{eq:reference_correction} uses the frozen estimate
$\hat z_i^{\mathrm{ref}}$ for the sampled flow time and noise.
The compatibility coefficient is
$q_{ij}=\exp(-\delta_{ij}^2/3)=\exp(-d_{ij}/3)$, corresponding to
$\alpha_{ij}=q_{ij}$ and $T_{\mathrm{comp}}=3$ in
Eq.~\eqref{eq:reference_correction}.
It retains more of the frozen estimate when the selected reference is
less compatible.
Policy updates remain conditioned on the original observation $o_i$.
Reference selection is performed offline and is not invoked during
policy inference.

\subsection{Offline Update Objective and Optimization}
\label{app:offline_update_implementation}

\paragraph{Reference-guided endpoint correction.}
Let $v_0$ denote the frozen pre-improvement policy's velocity field.
For a failed prediction $\hat z_i$ and its selected reference $z_j^{\mathrm e}$,
we sample $s\sim\mathcal U(0.25,0.75)$ and Gaussian noise $\epsilon$,
and construct $x_s=(1-s)\epsilon+s\hat z_i$.
Under the failed window's original observation $o_i$, the frozen policy
defines a local endpoint and a target velocity:
\begin{equation}
\begin{aligned}
h_0&=x_s+(1-s)v_0(x_s,s,o_i),\\
v_{ij}^{*}&=v_0(x_s,s,o_i)
+\frac{q_{ij}(z_j^{\mathrm e}-h_0)}{1-s}.
\end{aligned}
\label{eq:implemented_repair_target}
\end{equation}
The global correction strength is one, and $q_{ij}$ scales the target
displacement. With $D=Kd_z$, the repair loss is
\begin{equation}
\mathcal L_{\mathrm{repair}}
=\mathbb E_{(i,j)\sim\mathcal P,\,s,\epsilon}
\left[\frac{1}{D}\left\|v_\theta(x_s,s,o_i)
-\operatorname{sg}(v_{ij}^{*})\right\|_F^2\right].
\label{eq:implemented_repair_loss}
\end{equation}
There is no additional multiplication of this loss by $q_{ij}$.
The equivalent endpoint target is
$z_{ij}^{*}=(1-q_{ij})h_0+q_{ij}z_j^{\mathrm e}$.
Thus, the endpoint-error form in Eq.~\eqref{eq:offline_policy_improvement}
uses $\lambda_{\mathrm{corr}}=1$ and
$\lambda(s)=1/[D(1-s)^2]$ to match the implemented velocity MSE.

\paragraph{Optimization.}
We perform 2,000 policy updates using AdamW with weight decay $10^{-4}$
and global gradient-norm clipping at 1.0.
The learning rate warms up to $2\times10^{-5}$ over 50 updates, then
follows a cosine decay to $2\times10^{-6}$ by the end of training.

\FloatBarrier

\section{Experimental Details and Additional Results}
\label{app:experimental_details}

\subsection{Data Collection and Task Procedures}
\label{task}

We detail the data and task procedures used in Sec.~\ref{sec:setup}.

\paragraph{Demonstration collection.}
We collect 150 expert demonstrations for each of the four tasks,
yielding 600 demonstrations in total, by physically guiding the robot arms.
Each demonstration contains external and wrist-camera RGB observations,
fingertip tactile deformation fields, robot proprioception, and actions,
synchronized at 30\,Hz.

\paragraph{Data preprocessing.}
After collection, we apply image inpainting to the top-view RGB images
to remove the operator's arms and hands. The processed images are then
used for model training.

\paragraph{Rollout data for offline improvement.}
For each task, we select 50 successful rollout
trajectories for reward-weighted reinforcement and 25 failed
rollout trajectories for reference-guided retargeting.
The successful trajectories provide the interaction chunks in
$\mathcal D^+$, while low-reward chunks from the selected failed
trajectories serve as candidate sources for constructing the
reference pairs $\mathcal P$.
Only failed chunks that satisfy the reward and reference-selection
criteria described in Appendix~\ref{app:reference_selection}
contribute to the retargeting objective.

\begin{figure}[t]
  \centering
  \includegraphics[width=\linewidth]{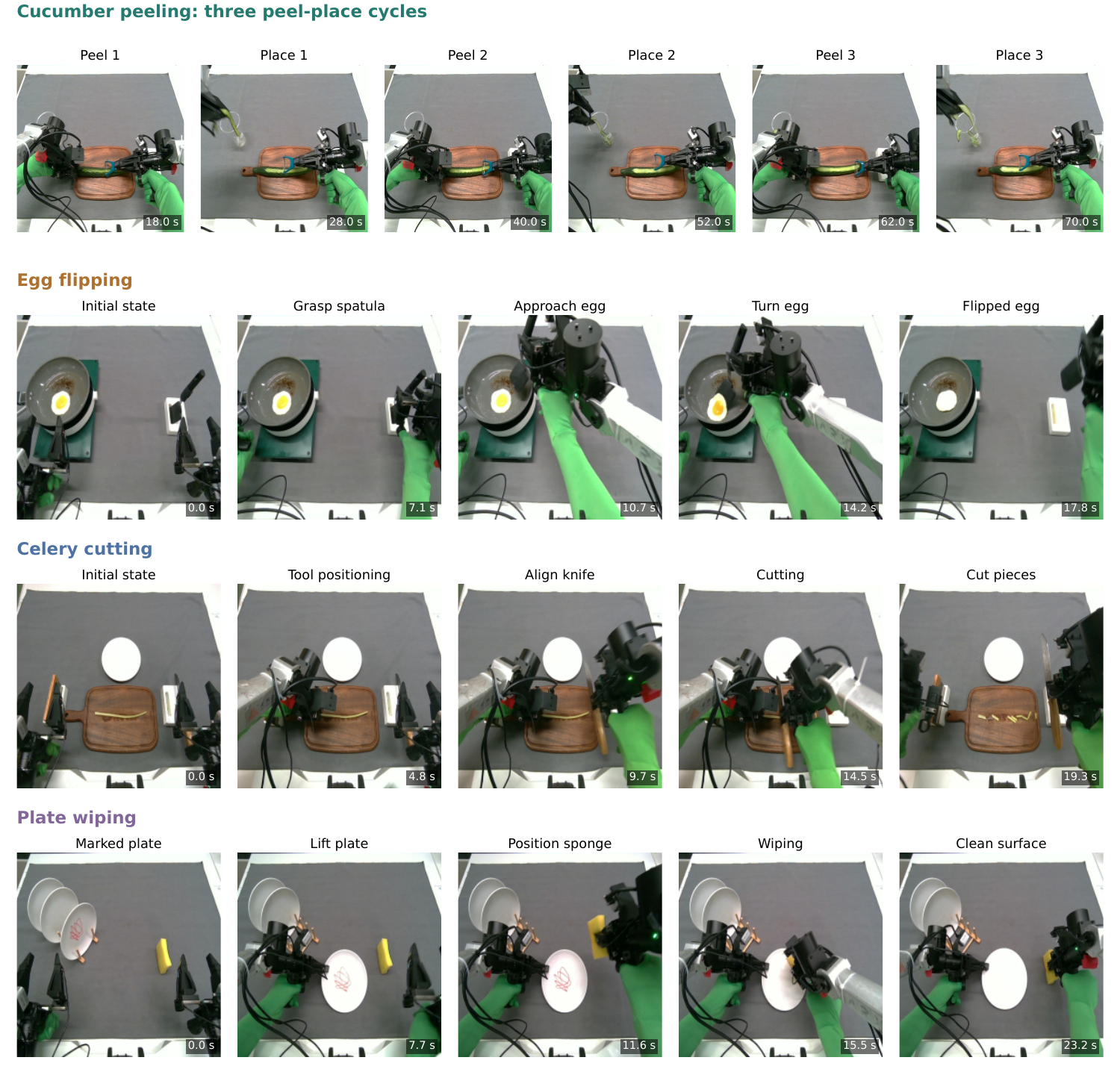}
  \caption{\textbf{Real-world task procedures.} Frames in each row are drawn
  from a single demonstration and ordered chronologically from left to right.
  Cucumber peeling comprises three successive peel--place cycles, shown as
  six alternating stages. The other rows illustrate tool preparation,
  object interaction, and the resulting state for flipping, cutting, and
  wiping. Timestamps are relative to the beginning of each recorded episode.}
  \label{fig:task_procedures}
\end{figure}

\paragraph{Task procedures.}
Fig.~\ref{fig:task_procedures} illustrates the four tasks using chronological
frames from recorded demonstrations.

\begin{itemize}
\item \textbf{Cucumber peeling.}
Each rollout consists of three consecutive peel--place cycles. In each
cycle, the robot brings a peeler into contact with the cucumber, moves it
along the surface to remove a complete, thick strip, and grasps and places
the removed strip before starting the next cycle.

\item \textbf{Egg flipping.}
The robot grasps a spatula, approaches the egg in the pan, slides the spatula
underneath it, and turns it over.

\item \textbf{Celery cutting.}
The robot aligns a knife with the celery and performs successive cutting
strokes. Each rollout contains multiple cutting attempts.

\item \textbf{Plate wiping.}
The robot grasps a marked plate and a sponge, establishes sponge--plate
contact, and wipes the marked surface while holding the plate.
\end{itemize}

\subsection{Evaluation Metrics}
\label{app:evaluation_metrics}

We define the success and contact-error metrics used in
Sec.~\ref{sec:setup} and Table~\ref{tab:ila_main_results}.

\paragraph{Success rate (SR).}
For the main robot experiments, each method is evaluated over 20 rollouts per task.
We report $\mathrm{SR}=100\times N_{\mathrm{success}}/N_{\mathrm{attempts}}$.
An attempt is a complete rollout for peeling, flipping, and wiping, and an
individual cut for celery cutting. Each cutting rollout contains six or
seven attempts; cutting SR pools successful and attempted cuts across
rollouts. Success is determined by the task
criteria below, independently of the tactile prediction errors.

\begin{itemize}
\item \textbf{Cucumber peeling.}
A rollout succeeds only when all three peel--place cycles are completed,
with a complete, thick strip removed, grasped, and placed in each cycle.
Insufficient tool engagement, a broken strip, and failure to grasp or place
the removed strip count as failures. A visibly peeled surface alone does
not satisfy the completion criterion.

\item \textbf{Egg flipping.}
A rollout succeeds when the egg is flipped over in the pan. Moving the
spatula over the egg without supporting it, or inserting it too shallowly
to complete the flip, counts as failure.

\item \textbf{Celery cutting.}
Each attempted cut is successful only if the celery is completely
severed at the cutting location before the knife clears the stalk.
Residual fibers that keep the pieces connected, including cases
where the partially cut stalk follows the lifted knife, count as failures.

\item \textbf{Plate wiping.}
A rollout succeeds only if all marks are removed. Partial wiping and any
remaining marks count as failure; simply moving the sponge across the
plate is insufficient.
\end{itemize}

\paragraph{Tactile fields and evaluation units.}
Each of the four fingertip sensors records a $35\times20$ field of
three-dimensional deformations $(d_x,d_y,d_z)$.
For prediction evaluation, let $j$ index a sensor at an evaluated target
frame, and let $S_j(p)\in\mathbb R^3$ and $\widehat S_j(p)$ denote its recorded
and predicted deformations at taxel $p\in\Omega$, where $|\Omega|=700$.
Predictions are evaluated after reversing training normalization, in the
original Xense Mesh3DFlow deformation units.
The four sensors are evaluated in their respective local coordinate systems.

\paragraph{Contact-region tactile error (Tac.\ L1).}
We define the contact mask from the recorded field as
$M_j(p)=\mathbf{1}[|S_j^z(p)|>0.005]$ and let $n_j=\sum_{p\in\Omega}M_j(p)$.
For each sensor--frame with $n_j>0$, we average absolute errors over its
contact taxels and all three deformation components, then average these
errors equally across the valid sensor--frames:
\[
e_j^{\mathrm{tac}}
=\frac{\sum_{p\in\Omega}M_j(p)
\|\widehat S_j(p)-S_j(p)\|_1}{3n_j},
\qquad
\mathrm{Tac.\ L1}
=\frac{1}{|\mathcal J_{\mathrm c}|}
\sum_{j\in\mathcal J_{\mathrm c}}e_j^{\mathrm{tac}},
\]
where $\mathcal J_{\mathrm c}=\{j:n_j>0\}$.
Sensor--frames with no recorded contact are excluded from this metric.
Thus, each valid sensor--frame has equal weight regardless of its contact
area, and prediction errors outside the recorded contact region are not scored.
The contact threshold defines the evaluation mask; task success follows
the manipulation criteria above.

\paragraph{Force proxy and prediction error ($F_{\mathrm{pred}}$ L1).}
For each sensor--frame, we sum the signed tangential deformations over
all taxels and take the magnitude of the resulting vector:
\[
g(S_j)=\left\|
\sum_{p\in\Omega}
\begin{pmatrix}S_j^x(p)\\S_j^y(p)\end{pmatrix}
\right\|_2,
\qquad
F_{\mathrm{pred}}\mathrm{\ L1}
=\frac{1}{|\mathcal J|}\sum_{j\in\mathcal J}
\left|g(\widehat S_j)-g(S_j)\right|.
\]
Here $\mathcal J$ contains all evaluated sensor--frames, including those
without recorded contact. The spatial sum uses all 700 taxels without a
contact mask. Resultants are computed separately for each sensor, and
their scalar magnitude errors are pooled; vectors from different sensors
are never summed together.
This quantity is a tangential deformation proxy for force application
in sensor-native units, with no conversion to newtons or training-time RMS scaling.
For the open-loop comparison, the proxy is computed from the predicted
tactile field rather than the separate aggregate-deformation head.

\paragraph{Execution force-proxy error ($F_{\mathrm{ext}}$ L1).}
For each task, we construct an expert reference by averaging the force-proxy
profiles of ten demonstrations collected using the operator's preferred
execution strategy.
We manually annotate the boundaries of task-required force-application
intervals in the expert demonstrations and robot rollouts.
Corresponding annotated intervals are matched for evaluation,
excluding phases in which force application is not required.
The evaluated intervals are defined by the required task interactions;
insufficient or absent force within such an interval contributes to the error.
Each rollout contributes one error value, and the reported mean $\pm$
standard deviation is computed across these trajectory-level errors.
Thus, $F_{\mathrm{ext}}$ L1 evaluates realized interaction during task-relevant
phases, while $F_{\mathrm{pred}}$ L1 evaluates predicted contact against recorded
future contact. Both complement SR, since agreement with a force-proxy
profile alone does not establish task completion.

\paragraph{Prediction-metric aggregation.}
The prediction evaluator pools errors over the sensor--frame instances
defined above. When dispersion is reported, it uses the population
standard deviation of those errors, rather than a standard deviation
across training seeds or trajectory means.
Table~\ref{tab:ila_open_loop_prediction} reports the mean prediction errors
for each task; its Average columns are unweighted means of the four task
values. Lower L1 values indicate better agreement with the corresponding
targets.

\subsection{Offline Policy-Improvement Baselines}
\label{app:offline_baselines}
We detail the baseline adaptations evaluated in Sec.~\ref{sec:offline_improvement_eval}.
All methods use the same frozen ILA representation and fixed rollout corpus.
RECAP initializes its flow-matching actor from the imitation-trained FM-ILA
policy. For IQL, we fit the Gaussian actor on successful rollout data before
offline RL updates. The methods otherwise retain their respective actor
parameterizations, reward formulations, and improvement objectives.
Both operate in the frozen interaction-latent space, with a normalized latent chunk $z_{\mathrm{normalized}}\in\mathbb{R}^{32\times128}$ as the policy output.
IQL uses a Gaussian actor and observation-conditioned critics $Q(o,z)$, updating the actor through advantage-weighted latent log-likelihood.
RECAP uses a flow-matching actor with positive and negative quality conditioning during training and positive conditioning at inference.
Both methods denormalize the predicted latent and pass it through the frozen Stage-1 decoder to obtain motor actions $a\in\mathbb{R}^{128\times14}$.
Thus, the baselines share the ILA action space and decoding interface while retaining their respective actor parameterizations and improvement objectives.

Both baselines receive 2,000 actor-improvement updates, with additional BC and critic/value fitting as specified in
Table~\ref{tab:offline_baseline_details}.
Training uses causal observations and task-outcome feedback, with nonoverlapping 128-frame transition chunks and valid physical durations retained for return computation.
These transitions follow the execution protocol in
Appendix~\ref{app:implementation_details}: the successor observation
is taken after the current action chunk has been executed in full.
For cutting, offline training uses episode-outcome labels, whereas task evaluation scores individual cuts.

\begin{table}[t]
\centering
\small
\setlength{\tabcolsep}{4pt}
\caption{\textbf{Offline RL baseline adaptations.} Each baseline receives
2,000 actor-improvement updates; additional BC and critic/value fitting
are specified below.}
\label{tab:offline_baseline_details}
\begin{tabularx}{\linewidth}{p{0.12\linewidth}X}
\toprule
Method & Key implementation settings \\
\midrule
IQL & Twin Q-functions and an expectile value function with $\tau=0.7$.
Terminal success receives $+1$; all other rewards are zero. The discount
is $0.99^{\Delta/128}$ for a transition of $\Delta$ physical frames.
The Gaussian actor has two width-256 Transformer decoder layers,
eight heads, and 32 latent queries; Q/V networks have two width-512
hidden layers. Actor log-likelihood weights are $\min\{\exp(3A),100\}$, where
$A=\min(Q_1,Q_2)-V$. Inference uses the conditional mean.
Additional training comprises 5,000 success-only BC updates and 5,000 critic
updates, followed by one critic update per actor-improvement update. \\
RECAP & A flow-matching actor uses positive/negative quality conditioning,
with 30\% condition dropout and a training-calibrated target positive
fraction of 40\%. A distributional value model with 201 bins on $[-1,0]$
is fitted to Monte Carlo returns for 5,000 updates. Advantage estimates
use one-chunk lookahead. Inference uses positive conditioning with
guidance scale 1. \\
\bottomrule
\end{tabularx}
\end{table}

RECAP uses a task-time reward: $-1$ per nonterminal physical frame,
zero at terminal success, and $-C_{\mathrm{fail}}$ at terminal failure.
We set $C_{\mathrm{fail}}=T_{\max}+1$, where $T_{\max}$ is the maximum
training episode length for the task, and divide all costs and returns
by the fixed scale $T_{\max}+C_{\mathrm{fail}}$.

\subsection{Reward-Design Comparisons}
\label{app:reward_design}
We detail the alternative rewards compared in Table~\ref{tab:ila_reward_ablation}.
Each variant replaces the reward used for
successful-sample weighting and failure/reference selection, while
keeping the policy initialization, observation interface, representation,
and update budget fixed. Reference pairs are rebuilt using the substituted
reward. These comparisons evaluate reward design within our procedure;
they do not reproduce complete external training systems.

\emph{Grasp-style contact reward.}
For sensor $a$, we construct a nonnegative contact map
$M^a=\max(-S_z^a,0)/q_{99}$, where $q_{99}$ is the 99th percentile of
training deformation samples, shared across sensors. Let $f^a$ denote
the spatial mean of $M^a$. The reward is
\begin{equation}
r^{\mathrm{grasp}}=
\frac{1}{|\mathcal A|}\sum_{a\in\mathcal A}
\operatorname{clip}\!\left(
\frac{f^a-f^a_{\mathrm{contact}}}
{f^a_{\max}-f^a_{\mathrm{contact}}},0,1\right),
\end{equation}
where $\mathcal A$ is the set of active sensors. The contact threshold
is 5\% of each sensor's P95 strength over training frames, and the upper
threshold is calibrated from P95 strength among contact frames.
This reward increases with contact strength and saturates at one;
it does not penalize excessive contact.

\emph{HapticVLA-style safety reward.}
We adapt the tactile safety terms of
HapticVLA~\citep{gubernatorov2026hapticvla} to our four deformation sensors.
The reward is the negative weighted sum of six penalties: excessive
contact strength, insufficient strength during holding, peak deformation,
spatial concentration, within-gripper asymmetry, and slip. Their weights
are $(1,1,1,1,1,0.1)$, respectively. The safe strength band uses P10/P95
of training contact frames, while peak and concentration limits use P99.
Holding begins after three consecutive contact frames with the gripper
closed and persists until the gripper opens. Slip is detected from a
contact-center jump or a sudden strength drop. Thresholds are fitted on
training demonstrations and then fixed. This adaptation uses the safety
reward only, without tactile distillation or episode-outcome reward terms
from the original method.

For both rule-based rewards, framewise scores are averaged over each
four-frame latent interval, and chunk scores average the resulting
32 token rewards. Scores are standardized using fixed training-set
statistics, with larger values indicating better interactions.

\subsection{Additional Motion--Contact Representation Analysis}
\label{app:representation_perturbations}

We detail the cucumber-peeling linear probes in
Sec.~\ref{sec:representation} and extend the input-perturbation analysis
to egg flipping, plate wiping, and celery cutting.

\paragraph{Probe data and splits.}
We partition the cucumber-peeling probe data by complete trajectory
into fitting, tuning, and test sets.
The test set contains ten held-out trajectories.
Fitting-chunk starts are separated by at least 16 frames, with at most
128 chunks per trajectory; tuning and test chunks are non-overlapping
within each trajectory. ILA and ILA w/o tactile use identical chunks
and targets. Tuning trajectories participated in Stage~1 training but
are excluded from probe fitting; test trajectories were used in neither
Stage~1 training nor probe fitting or tuning.

\paragraph{Linear readout and model selection.}
With each encoder frozen, we flatten its full $32\times128$ latent into
a 4,096-dimensional input and fit a separate multi-output ridge
regression with an intercept. Each target contains 128 frames of
14 raw action coordinates and eight signed deformation components:
the spatially summed $x$ and $y$ deformations of each of four sensors.
Inputs and target channels are standardized using fitting-set
statistics, and predictions are restored to their original scales
before evaluation. For standardized inputs $X$ and flattened targets
$Y$, the fitting objective is
\[
\min_{W,b}\;
\frac{1}{N}\left\|XW+\mathbf{1}b^{\mathsf T}-Y\right\|_F^2
+\lambda\|W\|_F^2,
\]
where $N$ is the number of fitting chunks.
For each representation, we select
$\lambda\in\{0.001,0.01,0.1,1,10\}$ by minimizing tuning-set standardized
MSE, with equal weight assigned to the action and contact groups.
Both full-dimensional probes select $\lambda=0.001$.
The selected probes retain the fitting-set coefficients, without
refitting on the combined fitting and tuning sets.

\paragraph{Readout metrics and uncertainty.}
We report test L1 for raw actions and force-proxy magnitudes.
For the latter, we take the two-dimensional norm of each sensor's
predicted signed deformation resultant and compare it with the
ground-truth norm. Thus, the magnitude is derived from the linear
vector prediction, in the sensor-native deformation units defined in
Appendix~\ref{app:evaluation_metrics}.
We first average each metric within each test trajectory, then average
the ten trajectory errors with equal weight.
For 95\% confidence intervals, we draw 2,000 bootstrap samples of ten
trajectories with replacement using seed 42 and take the 2.5th and
97.5th percentiles of the resulting means. The same resamples are used
for both representations. Encoders and probes remain fixed throughout
bootstrap resampling; the intervals quantify test-trajectory sampling
uncertainty conditional on these fitted models and exclude variation
across training seeds.

\paragraph{Scope of the probes.}
The probes assess motion and contact information recoverable from the
frozen interaction encoders' outputs for complete recorded chunks.
Future-contact prediction from policy observations is evaluated
separately in Appendix~\ref{app:contact_prediction}.

\paragraph{Controlled input perturbations.}
We use the frozen ILA encoder for each task and all complete, non-overlapping 128-step chunks from its held-out trajectories: 25 chunks from five egg-flipping trajectories, 30 from five plate-wiping trajectories, and 17 from two celery-cutting trajectories.
Each intervention modifies either actions or tactile deformation fields, while keeping the other modality and visual context fixed.
Action scaling follows $a'_\tau=a_0+\alpha(a_\tau-a_0)$, where $a_0$ is the chunk's initial motor command; tactile scaling follows $S'=\alpha S$.
We apply both interventions before the saved training normalization, using $\alpha\in\{1,0.75,0.5,0.25,0\}$.
Separately, we add elementwise Gaussian noise to normalized inputs, $x'=x+\sigma\epsilon$, with $\sigma\in\{0,0.1,0.2\}$ and independent standard-normal entries in $\epsilon$.
We use seeds 11, 22, and 33, reusing the same noise realization across nonzero levels for each chunk, modality, and seed without clipping.
Equal noise standard deviations do not imply equal total perturbation norms across modalities.

\begin{figure}[t]
    \centering
    \includegraphics[width=\linewidth]{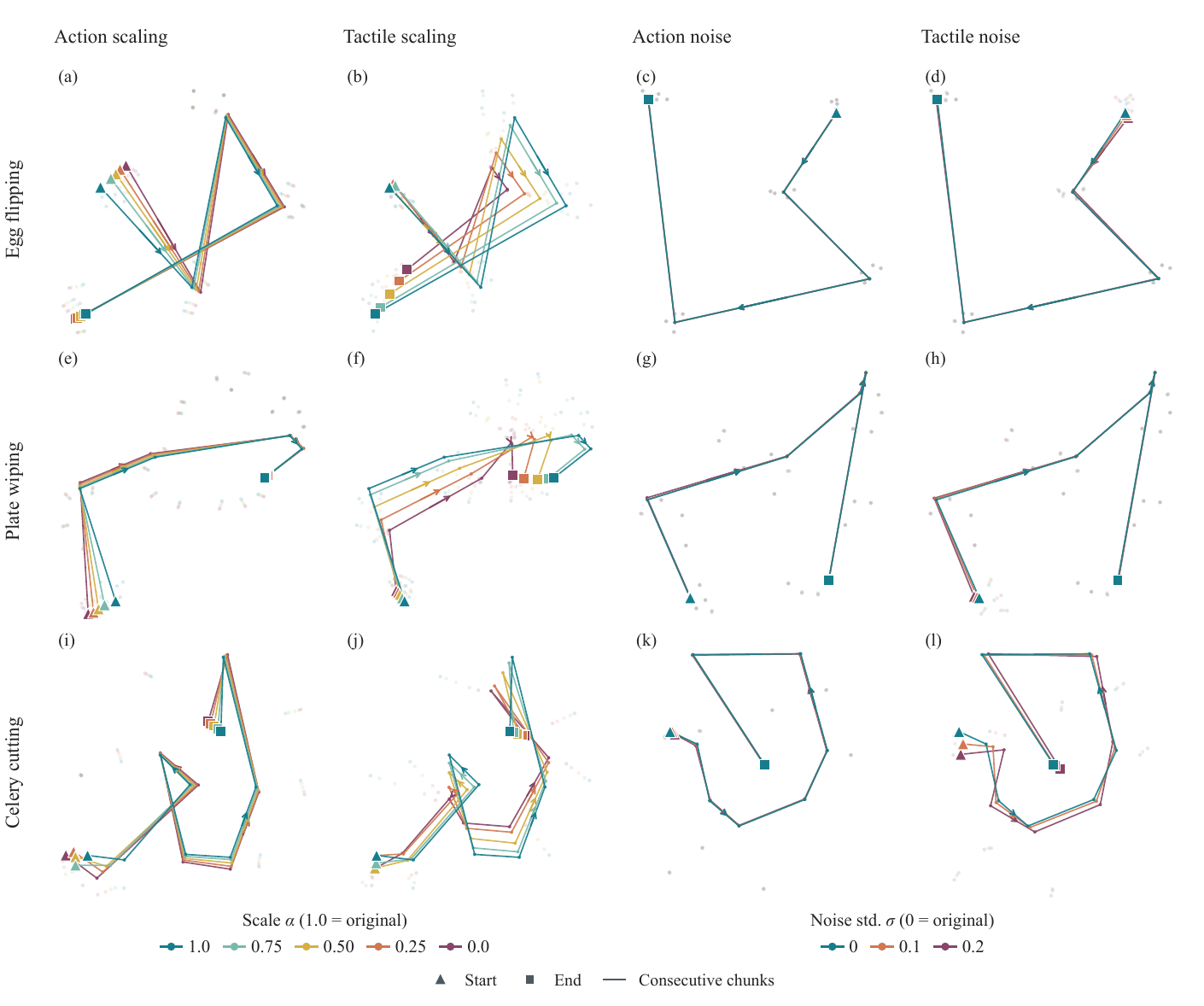}
    \caption{\textbf{Motion--contact representation analysis across additional tasks.}
    Rows show egg flipping, plate wiping, and celery cutting.
    Columns show action and tactile scaling, followed by Gaussian noise in normalized action and tactile inputs; other inputs remain fixed.
    Colored paths connect consecutive chunks of one held-out trajectory per task; faint points show all test chunks under the displayed conditions.
    Scaling and noise use separate t-SNE projections within each task.}
    \label{fig:ila_perturbations_supplement}
\end{figure}

\paragraph{Visualization protocol.}
We flatten each complete $32\times128$ latent, reduce it to 50 principal components, and apply t-SNE with perplexity 40, 1,500 iterations, and random seed 42.
Within each task, action and tactile scaling conditions share one projection, while noise conditions across both modalities and all seeds share another; the unperturbed baseline enters each fit only once.
Figure~\ref{fig:ila_perturbations_supplement} displays noise seed 11 and connects consecutive chunks from one trajectory per task, selected by the lower median of sorted trajectory IDs rather than by visual separation.
Faint points show all test chunks under the displayed conditions.
Coordinates are not directly comparable across tasks or between scaling and noise projections.

\paragraph{Motion and contact responses.}
Across all three tasks, tactile scaling changes the projected latent paths even when motor actions remain fixed, while action scaling produces more localized changes along the displayed paths.
At the tested noise levels, paths largely overlap, although tactile noise produces visible deviations on celery cutting.
These observations provide qualitative evidence of contact-sensitive action representations across tasks, complementing the cucumber-peeling linear probes in Sec.~\ref{sec:representation}.
They characterize encoder responses to synthetic perturbations, rather than establishing the physical feasibility of the modified interactions or robustness of closed-loop execution.

\subsection{Qualitative Visualization of Interaction Representations}
\label{app:representation_visualization}

\begin{figure}[t]
    \centering
    \includegraphics[width=0.85\linewidth]{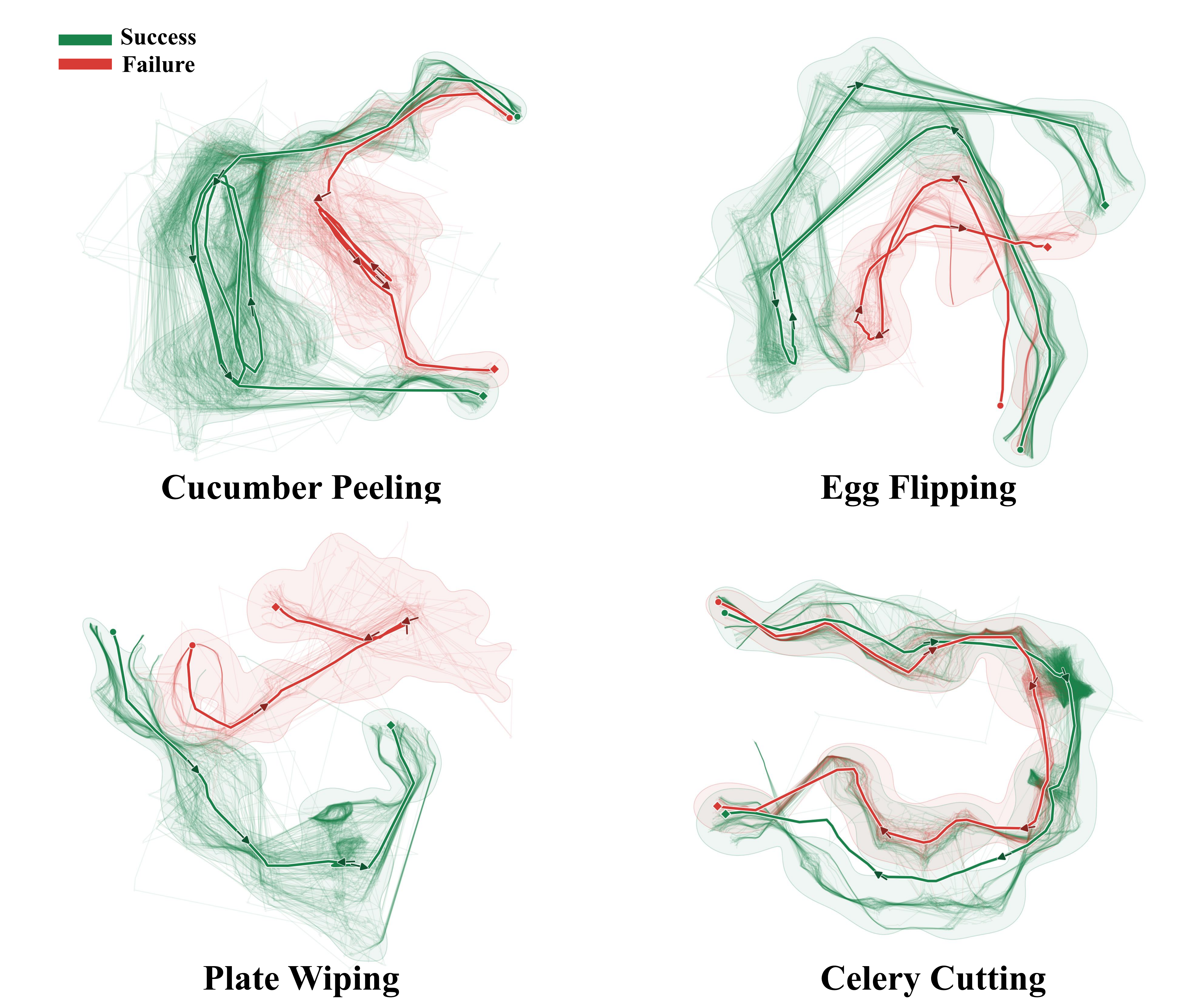}
    \vspace{-3mm}
    \caption{
    \textbf{Qualitative visualization of interaction representations.}
    Two-dimensional UMAP projections for cucumber peeling,
    egg flipping, plate wiping, and celery cutting.
    Green and red denote representations from successful and
    failed episodes, respectively.
    Lines connect representations in temporal order.
    Shared trajectory regions and local deviations illustrate
    variation across recorded interactions; apparent distances
    and separation should be interpreted qualitatively.
    }
    \vspace{-2mm}
    \label{fig:interaction_representation}
\end{figure}

Fig.~\ref{fig:interaction_representation} visualizes
two-dimensional UMAP projections of interaction representations
for four manipulation tasks, complementing Sec.~\ref{sec:representation}.
Success and failure trajectories
are colored by their episode-level outcomes. These outcome labels
describe the complete episode; they do not imply that every
representation within a failed episode corresponds to an
unsuccessful interaction.

The visualization reveals both shared and outcome-dependent
trajectory structure. Successful and failed episodes overlap
along parts of their trajectories, while deviations emerge in
other regions. Such overlap is expected: episodes with different
final outcomes can share similar motions and contact conditions
during portions of execution. The observed deviations are
consistent with differences in interaction dynamics, but their
physical interpretation requires inspection of the corresponding
observations.

We use this visualization as a qualitative diagnostic rather
than a quantitative measure of representation quality.
Two-dimensional UMAP projections can distort distances and
global geometry, and apparent separation does not establish
separability in the original representation space.
Likewise, overlap does not imply that the underlying
representations are identical. The plots alone do not establish
that the representation isolates task outcome from task phase,
motion, or other correlated factors, nor do they validate
reward calibration. Quantitative evaluation is therefore needed
to assess the utility of the learned representation.

\subsection{Open-Loop Tactile and Force-Proxy Prediction}
\label{app:contact_prediction}

We evaluate whether policy-generated interaction latents retain predictive information about future contact.
This complements the robot experiments in Sec.~\ref{sec:main_results} and the encoder analysis in Sec.~\ref{sec:representation} by assessing the tactile and force-related content decoded from policy predictions.

\paragraph{Prediction setting.}
We evaluate on ten additional expert trajectories per task (40 in total),
held out from training. Evaluation covers each trajectory from the start
to the end of task execution, with all segmented interaction chunks
included in prediction and metric aggregation.
Each policy predicts future interactions from the observations available
at the start of each chunk.
We use the final policy checkpoints at the end of the training schedules
specified in Appendix~\ref{app:implementation_details}, with the same
checkpoint used for both prediction metrics.
For ILA, we decode tactile deformation fields from the policy-generated latent using the frozen decoder and compute force proxies from these fields.
The latent is predicted from observations, rather than obtained by encoding the ground-truth future action--tactile sequence.
We compare FM-ILA and VLA-ILA with their corresponding P-tac$^{*}$ variants, which jointly predict motor actions and tactile signals, and WAM-ILA with DreamTac~\citep{lou2026dream}.
Tac.\ L1 and $F_{\mathrm{pred}}$ L1 follow the contact-region and force-proxy definitions in Appendix~\ref{app:evaluation_metrics}.
We report each metric per task and its unweighted mean across the four tasks.

\begin{table}[t]
\centering
\caption{\textbf{Open-loop tactile and force-proxy prediction.}
We report mean Tac.\ L1 and $F_{\mathrm{pred}}$ L1; Average is the unweighted mean across the four tasks.}
\label{tab:ila_open_loop_prediction}
\setlength{\tabcolsep}{3pt}
\renewcommand{\arraystretch}{1.1}
\resizebox{\linewidth}{!}{%
\begin{tabular}{llcccccccccc}
\toprule
 & & \multicolumn{2}{c}{Cucumber Peeling} & \multicolumn{2}{c}{Egg Flipping} & \multicolumn{2}{c}{Celery Cutting} & \multicolumn{2}{c}{Plate Wiping} & \multicolumn{2}{c}{Average} \\
\cmidrule(lr){3-4}\cmidrule(lr){5-6}\cmidrule(lr){7-8}\cmidrule(lr){9-10}\cmidrule(lr){11-12}
Family & Method & Tac.\ L1 $\downarrow$ & $F_{\mathrm{pred}}$ L1 $\downarrow$ & Tac.\ L1 $\downarrow$ & $F_{\mathrm{pred}}$ L1 $\downarrow$ & Tac.\ L1 $\downarrow$ & $F_{\mathrm{pred}}$ L1 $\downarrow$ & Tac.\ L1 $\downarrow$ & $F_{\mathrm{pred}}$ L1 $\downarrow$ & Tac.\ L1 $\downarrow$ & $F_{\mathrm{pred}}$ L1 $\downarrow$ \\
\midrule
FM & FM-tac$^{*}$ & $0.0125$ & $3.07$ & $0.0130$ & $4.16$ & $0.0083$ & $1.93$ & $0.0116$ & $2.61$ & $0.0114$ & $2.94$ \\
 & \textbf{FM-ILA} & $0.0124$ & $2.70$ & $0.0100$ & $2.57$ & $0.0068$ & $1.21$ & $0.0105$ & $1.48$ & $0.0099$ & $1.99$ \\
\cmidrule(lr){1-12}
VLA ($\pi_{0.5}$) & VLA-tac$^{*}$ & $0.0114$ & $2.81$ & $0.0128$ & $3.45$ & $0.0084$ & $2.26$ & $0.0111$ & $2.69$ & $0.0109$ & $2.80$ \\
 & \textbf{VLA-ILA} & $0.0122$ & $2.63$ & $0.0101$ & $2.45$ & $0.0070$ & $1.26$ & $0.0103$ & $1.37$ & $0.0099$ & $1.93$ \\
\cmidrule(lr){1-12}
WAM & DreamTac & $0.0094$ & $2.72$ & $0.0114$ & $3.29$ & $0.0061$ & $1.59$ & $0.0085$ & $2.11$ & $0.0088$ & $2.43$ \\
 & \textbf{WAM-ILA} & $0.0124$ & $2.78$ & $0.0100$ & $2.34$ & $0.0069$ & $1.23$ & $0.0111$ & $1.58$ & $0.0101$ & $1.98$ \\
\bottomrule
\end{tabular}%
}
\end{table}

\paragraph{Prediction accuracy.}
Table~\ref{tab:ila_open_loop_prediction} shows that ILA lowers both average prediction errors relative to explicit joint action--tactile prediction under FM and VLA backbones.
Mean $F_{\mathrm{pred}}$ L1 decreases from 2.94 to 1.99 for FM and from 2.80 to 1.93 for VLA, with improvements on all four tasks.
Mean Tac.\ L1 decreases from 0.0114 to 0.0099 and from 0.0109 to 0.0099, respectively.

The ILA variants obtain mean Tac.\ L1 values of 0.0099--0.0101, approaching DreamTac's 0.0088, while achieving lower mean force-proxy errors of 1.93--1.99 versus 2.43.
Within the WAM comparison, WAM-ILA reduces force-proxy error on three of the four tasks, with peeling as the exception.
DreamTac retains the lowest average pointwise tactile error, showing that local deformation accuracy and aggregate force-proxy accuracy capture complementary aspects of contact prediction.

\paragraph{Contact prediction through the action representation.}
DreamTac explicitly includes future tactile observations in its diffusion targets, jointly denoising them with visual and action tokens through a shared DiT~\citep{lou2026dream}.
ILA obtains contact predictions from its action latent through a lightweight frozen decoder, without additional denoising.
These results support the same latent representation for executable action generation and future contact prediction.

\subsection{Comparison with Action-Trajectory Replay}
\label{app:trajectory_replay}

We examine whether replaying the motor actions of a successful demonstration can reproduce its interaction outcome when object geometry and placement are approximately restored.
We compare this replay strategy with ILA policies to assess how reliably recorded commands and learned policies reproduce contact-rich behavior.

\paragraph{Replay protocol.}
After collecting a demonstration, we place the task objects at their demonstration positions and replay the recorded motor action sequence without adapting it to online visual or tactile observations.
For cucumber peeling and celery cutting, we select replacement objects with similar shapes and sizes no smaller than those used in the demonstration, so that the recorded trajectory retains an opportunity to make contact.
For egg flipping and plate wiping, we reuse the same egg and plate props.
This procedure approximately reproduces the spatial arrangement, while differences in object geometry can still affect contact location, penetration depth, and loading.
We evaluate replay over 20 rollouts per task.
We report task success rate (SR) and execution force-proxy error ($F_{\mathrm{ext}}$ L1, mean $\pm$ standard deviation) relative to expert reference profiles, following the metric definitions in Appendix~\ref{app:evaluation_metrics}.

\paragraph{Policy comparison.}
We reproduce the imitation-trained FM-ILA, VLA-ILA, and WAM-ILA results, together with FM-ILA after offline improvement, from Table~\ref{tab:ila_main_results}.
These policies generate interaction latents from current observations and decode them into motor actions.
Replay uses the object-matching protocol above, while the policy results follow the main evaluation protocol; this comparison therefore complements the controlled action-formulation comparisons in Sec.~\ref{sec:main_results}.

\begin{table}[t]
\centering
\caption{\textbf{ILA policies and action-trajectory replay.}
We report SR and $F_{\mathrm{ext}}$ L1 (mean $\pm$ standard deviation).}
\label{tab:ila_trajectory_replay}
\setlength{\tabcolsep}{3pt}
\renewcommand{\arraystretch}{1.1}
\resizebox{\linewidth}{!}{%
\begin{tabular}{lcccccccc}
\toprule
 & \multicolumn{2}{c}{Cucumber Peeling} & \multicolumn{2}{c}{Egg Flipping} & \multicolumn{2}{c}{Celery Cutting} & \multicolumn{2}{c}{Plate Wiping} \\
\cmidrule(lr){2-3}\cmidrule(lr){4-5}\cmidrule(lr){6-7}\cmidrule(lr){8-9}
Method & SR $\uparrow$ & $F_{\mathrm{ext}}$ L1 $\downarrow$ & SR $\uparrow$ & $F_{\mathrm{ext}}$ L1 $\downarrow$ & SR $\uparrow$ & $F_{\mathrm{ext}}$ L1 $\downarrow$ & SR $\uparrow$ & $F_{\mathrm{ext}}$ L1 $\downarrow$ \\
\midrule
Action-trajectory replay & 40\% & $1.11 \pm 0.55$ & 100\% & $1.49 \pm 0.79$ & 46\% & $6.54 \pm 3.30$ & 50\% & $4.05 \pm 0.32$ \\
\midrule
\multicolumn{9}{l}{\emph{Imitation-trained ILA policies}} \\
FM-ILA & 75\% & $2.04 \pm 0.37$ & 95\% & $0.56 \pm 0.64$ & 50\% & $7.52 \pm 0.78$ & 100\% & $3.44 \pm 0.35$ \\
VLA-ILA & 50\% & $2.28 \pm 0.41$ & 85\% & $0.38 \pm 0.54$ & 37\% & $7.22 \pm 1.17$ & 95\% & $2.51 \pm 0.38$ \\
WAM-ILA & 70\% & $1.44 \pm 0.47$ & 80\% & $0.14 \pm 0.50$ & 45\% & $8.94 \pm 0.63$ & 100\% & $1.33 \pm 0.49$ \\
\midrule
\multicolumn{9}{l}{\emph{After offline policy improvement}} \\
FM-ILA + update & 85\% & $0.51 \pm 0.50$ & 100\% & $0.49 \pm 0.55$ & 56\% & $5.55 \pm 0.38$ & 100\% & $0.72 \pm 0.34$ \\
\bottomrule
\end{tabular}
}
\end{table}

\paragraph{Results and interpretation.}
Table~\ref{tab:ila_trajectory_replay} shows that the difference between replay and ILA policies depends on the task.
On peeling and wiping, replay reaches 40\% and 50\% SR, compared with 75\% and 100\% for imitation-trained FM-ILA and 85\% and 100\% after offline improvement.
VLA-ILA and WAM-ILA also exceed replay SR on both tasks.
This pattern is consistent with the need to sustain tool--surface engagement, although the experiment does not isolate sustained contact from object variation and other task mechanics.
Replay achieves 100\% SR on egg flipping, matching improved FM-ILA.
On cutting, its 46\% SR falls within the 37--50\% range of imitation-trained ILA policies, whereas improved FM-ILA reaches 56\%.

Force-proxy error captures a complementary aspect of execution.
On peeling, replay has lower error than imitation-trained FM-ILA (1.11 versus 2.04) despite lower SR (40\% versus 75\%), illustrating that lower average profile error need not imply task completion.
After improvement, FM-ILA achieves higher SR than replay on peeling, cutting, and wiping, matches its flipping SR, and has lower force-proxy error on all four tasks; on cutting, its error is 5.55 versus 6.54 for replay.
Thus, recorded commands can reproduce some outcomes under approximately restored object conditions, but do not consistently reproduce either task success or expert contact profiles.
These results motivate representing contact evolution alongside motion.
The controlled policy comparisons and component ablations in Secs.~\ref{sec:main_results} and~\ref{sec:ablations} assess the specific contribution of ILA.

\subsection{Diagnostics of Offline Policy Improvement}
\label{app:offline_diagnostics}

We visualize the weighting and reference-selection mechanisms used in
Sec.~\ref{sec:interaction_improvement} and evaluated in
Sec.~\ref{sec:offline_improvement_eval}.

\paragraph{Successful-window reweighting.}

\begin{figure}[t]
    \centering
    \includegraphics[width=0.9\linewidth]
    {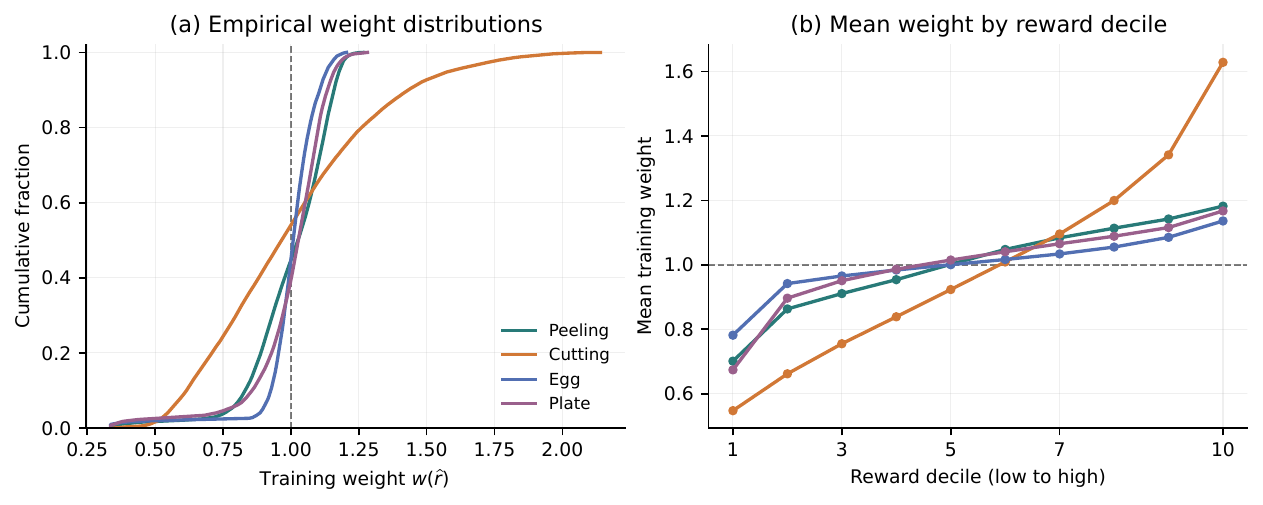}
    \caption{\textbf{Successful-window reweighting across four tasks.}
    Left: empirical weight distributions.
    Right: mean weight within each reward decile.
    Dashed lines indicate uniform weighting, $w_i=1$.
    Weights are recomputed from cached rewards and the fixed calibration
    of each completed run, using $\eta=0.3$, $\beta=2$, and
    $a_{\mathrm{clip}}=3$.
    The diagnostic pools contain successful-rollout policy predictions.}
    \label{fig:success_weight_distribution}
\end{figure}

Fig.~\ref{fig:success_weight_distribution} shows the weighting
rule applied to the successful training-window pools of four
completed runs.
The empirical distributions show the spread of the resulting
coefficients, and the reward-decile means show their dependence on
the cached scores.
The highest reward decile receives $1.69$, $2.98$, $1.45$, and
$1.73$ times the mean weight of the lowest decile for Peeling,
Cutting, Egg, and Plate, respectively.
Under uniform window sampling, these differences translate directly
into different coefficients in the training objective.

\paragraph{Observation-compatible reference pairs.}

\begin{figure}[t]
    \centering
    \includegraphics[width=\linewidth]{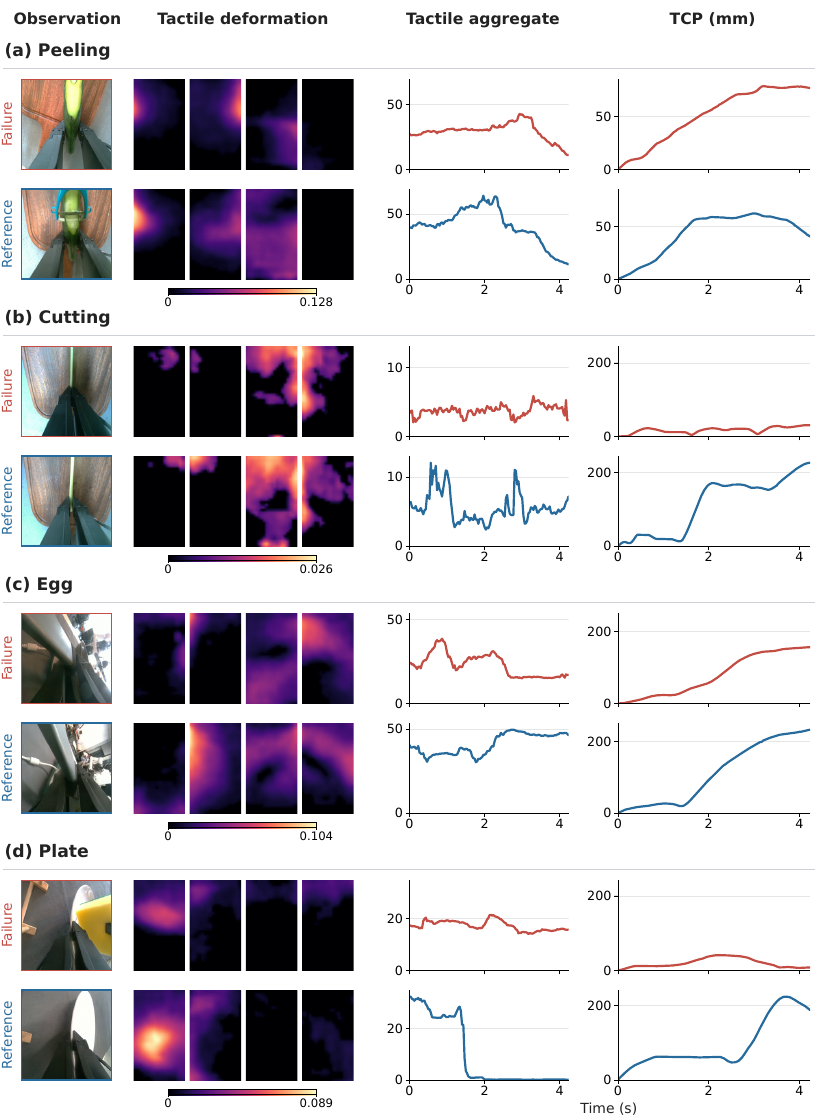}
    \caption{\textbf{Observation-compatible pairs across four tasks.}
    Failure and reference data occupy the upper and lower rows of each task.
    Wrist images and tactile maps are taken at the respective window starts.
    The four tactile maps show local tangential deformation magnitudes,
    with a shared color scale within each task.
    The tactile curve sums the four per-sensor spatial XY resultant
    magnitudes in raw deformation units.
    TCP shows the mean positional displacement of the two recorded
    end effectors from their respective window starts.}
    \label{fig:paired_tasks_compact}
\end{figure}

Figure~\ref{fig:paired_tasks_compact} shows four saved examples from
the archived recorded-window retrieval implementation.
The observations and continuations come from the stored source and
reference windows; the figure makes their correspondence inspectable
without treating the reference continuation as a corrected-policy
execution.

\FloatBarrier

\section{Towards a Shared Interaction Latent Space}
\label{app:shared_interaction_latents}

We additionally explore whether the interaction representation in
Sec.~\ref{sec:interaction_representation} can be learned jointly across tasks, rather than separately
for each task. Specifically, we pool demonstrations from all four tasks and
train a single interaction encoder and decoder with shared reconstruction heads.
The shared model is trained for approximately 100,000 optimizer steps.

\begin{figure}[h]
    \centering
    \includegraphics[width=0.5\linewidth]{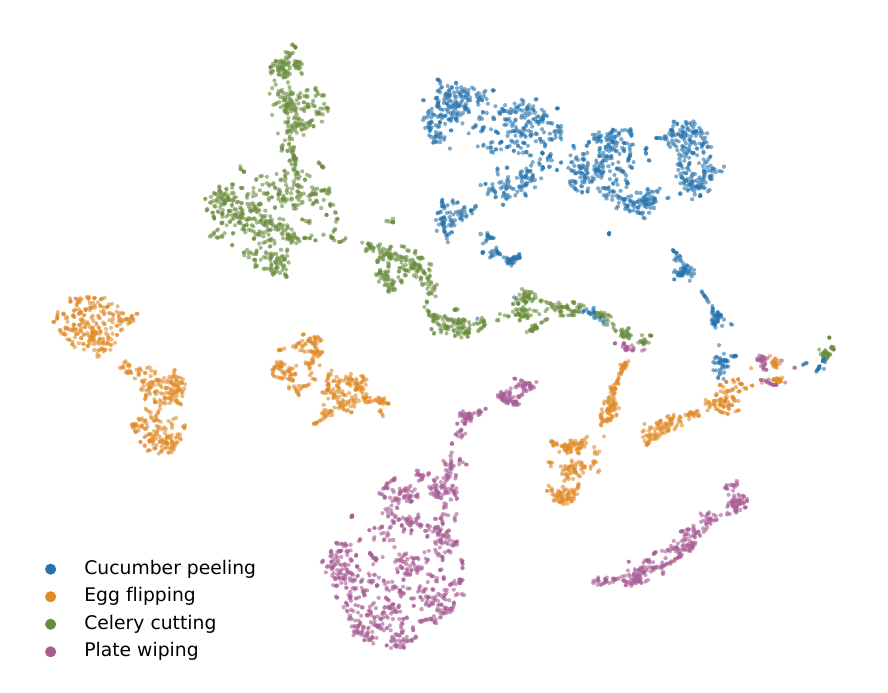}
    \caption{\textbf{Four tasks encoded by a shared Stage-1 model.}
    Each point represents a 128-dimensional token and is colored by task. All tasks are encoded by
    the same model with shared preprocessing.}
    \label{fig:shared_stage1_latents}
\end{figure}

\paragraph{Shared latent visualization.}
Figure~\ref{fig:shared_stage1_latents} visualizes all four tasks under the
same frozen encoder using a single t-SNE embedding. We sample 1,500 latent
tokens per task from successful demonstrations, approximately balancing
samples across available episodes. 
The resulting embedding exhibits task-associated structure within the shared
latent space. We emphasize that this visualization is descriptive and does
not by itself establish cross-task semantic alignment.

\paragraph{Reconstruction comparison.}
Table~\ref{tab:shared_stage1_reconstruction} compares the shared model with
task-specific pretraining on the same configured held-out episode sets.
The shared model
achieves lower action reconstruction error on all four tasks, while the
task-specific models achieve slightly lower pointwise tactile error.
Force-proxy reconstruction is lower with the shared model on peeling,
flipping, and cutting, but lower with the task-specific model on wiping.
Thus, pooling data across the observed tasks does not uniformly improve all
reconstruction quantities, but the shared encoder can represent all four
interaction domains without degrading action reconstruction.

\begin{table}[h]
\centering
\caption{\textbf{Shared versus task-specific reconstruction.}
Action and tactile report L1 errors in the original
recorded coordinates. Force reports magnitude L1 of the directly decoded
aggregate deformation, a force proxy in sensor-recording units rather than
newtons.}
\label{tab:shared_stage1_reconstruction}

\small
\setlength{\tabcolsep}{9pt}
\renewcommand{\arraystretch}{1.15}

\begin{tabular}{llccc}
\toprule
Task & Model
& Action $\downarrow$
& Tactile $\downarrow$
& Force $\downarrow$ \\
& & $\times 10^{-3}$ & $\times 10^{-3}$ & $\times 10^{-2}$ \\
\midrule

Cucumber peeling
& Task-specific & 34.21 & \textbf{3.50} & 25.80 \\
& Shared        & \textbf{22.63} & 3.51 & \textbf{23.72} \\
\addlinespace[3pt]

Egg flipping
& Task-specific & 18.05 & \textbf{2.39} & 25.37 \\
& Shared        & \textbf{6.12} & 2.87 & \textbf{16.54} \\
\addlinespace[3pt]

Celery cutting
& Task-specific & 6.90 & \textbf{1.21} & 14.25 \\
& Shared        & \textbf{6.55} & 1.44 & \textbf{8.29} \\
\addlinespace[3pt]

Plate wiping
& Task-specific & 6.28 & \textbf{2.22} & \textbf{2.18} \\
& Shared        & \textbf{4.35} & 2.54 & 7.23 \\

\bottomrule
\end{tabular}
\end{table}

\FloatBarrier

\clearpage
\section{Rollout Cases and Interaction Analysis}
\label{app:rollout_cases}

We examine recorded executions across all four tasks, first showing
successful interactions and then analyzing 17 selected failure cases.
The sequences expose distinct requirements for tool acquisition, sustained
contact, object support, and release. Frames are ordered chronologically
from left to right. These cases complement the
quantitative evaluation by showing where an apparently plausible motion
does or does not produce the required object state.

\begin{figure}[!htbp]
    \centering
    \includegraphics[width=\linewidth]{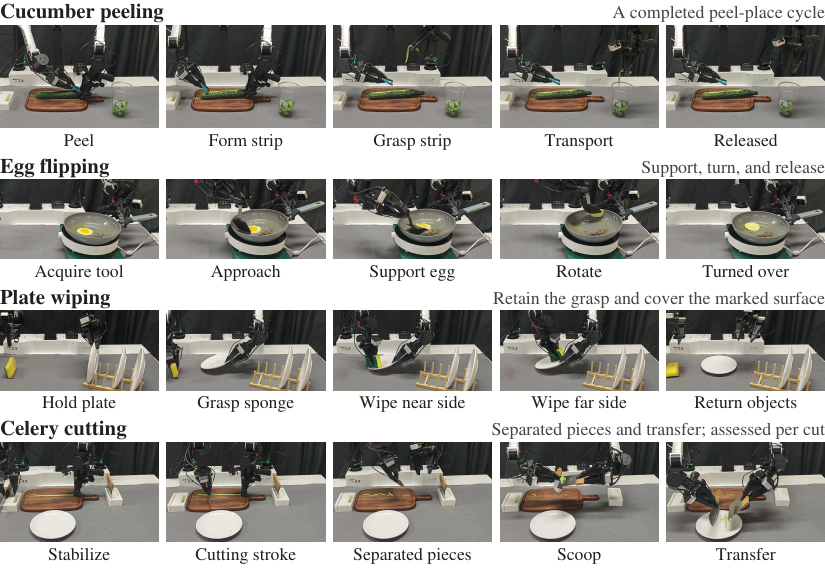}
    \caption{\textbf{Successful task interactions across four tasks.}
    The sequences show strip formation and transfer, egg turnover,
    plate wiping, and celery separation and transfer. The peeling row
    depicts one completed peel--place cycle, rather than all three cycles
    required for rollout success. Cutting is assessed per cut; the
    displayed separated pieces and transfer do not imply that every
    preceding cut succeeded.}
    \label{fig:rollout_success_interactions}
\end{figure}

\paragraph{Reading motion together with contact.}
Successful execution requires the object state to change with the
intended motion. For peeling, surface contact must first produce an
intact strip, followed by a separate grasp and release. Flipping requires
support under the egg before turning it. Wiping requires both retention
of the sponge and coverage of the marked surface. Cutting requires
stabilization and complete separation before the knife clears the stalk.
Table~\ref{tab:rollout_interaction_requirements} organizes these task-specific
requirements and the corresponding failure cases.

\begin{table}[!htbp]
\centering
\small
\setlength{\tabcolsep}{4pt}
\renewcommand{\arraystretch}{1.08}
\caption{\textbf{Interaction requirements illustrated by the rollout cases.}
Case IDs correspond to the rows of Figs.~\ref{fig:rollout_failures_peel_wipe}--\ref{fig:rollout_failures_cut}.}
\label{tab:rollout_interaction_requirements}
\begin{tabularx}{\linewidth}{@{}lXl@{}}
\toprule
Task & Required motion--contact coordination & Cases \\
\midrule
Peeling & Maintain a productive stroke; acquire and transfer the strip. & P1--P4 \\
Flipping & Support, turn, and release the egg while retaining the tool. & E1--E5 \\
Wiping & Retain both objects and move effective contact across the marks. & W1--W3 \\
Cutting & Stabilize, sever, and clear the knife between strokes. & C1--C5 \\
\bottomrule
\end{tabularx}
\end{table}

\begin{figure}[!htbp]
    \centering
    \includegraphics[width=0.97\linewidth]{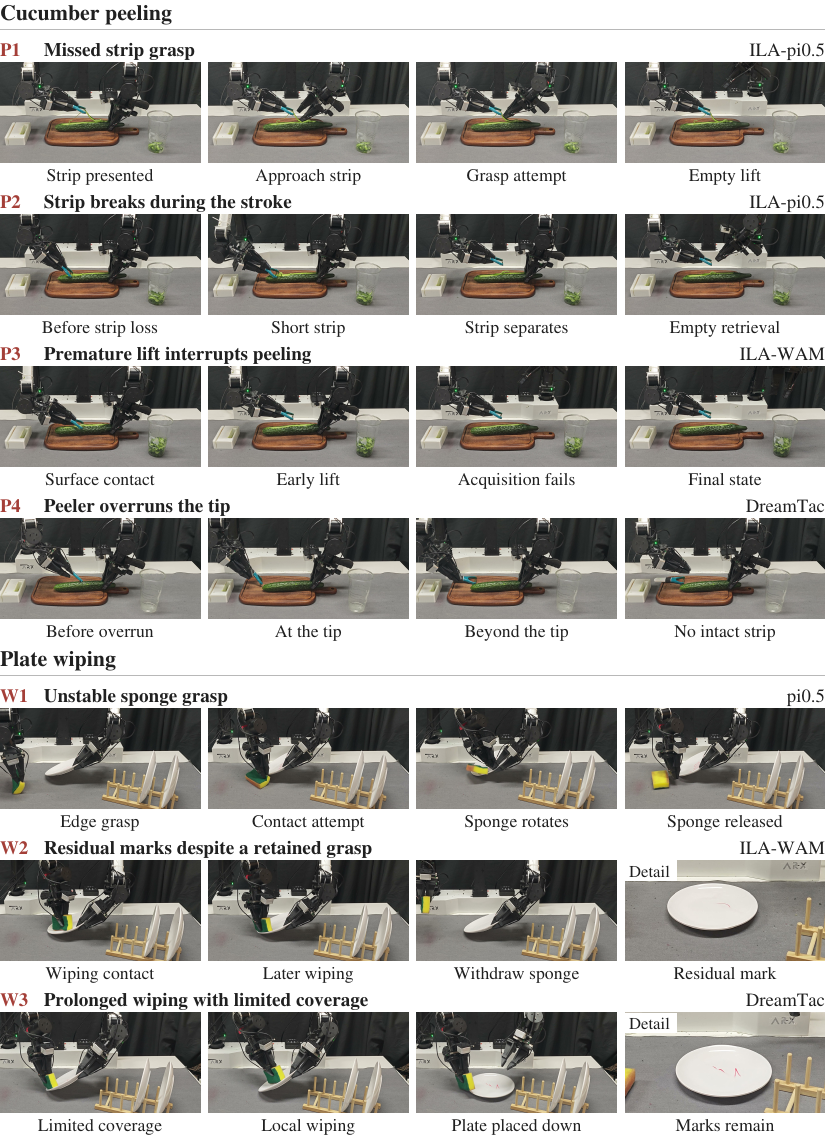}
    \caption{\textbf{Failure cases in cucumber peeling and plate wiping.}
    P1--P4 show missed acquisition, strip breakage, premature lifting,
    and tool overrun. W1--W3 show an unstable sponge grasp, residual
    staining despite a retained grasp, and prolonged wiping with limited
    coverage. W2--W3 end with enlarged views of residual markings.
    The images do not determine contact pressure.}
    \label{fig:rollout_failures_peel_wipe}
\end{figure}

\clearpage
\paragraph{Peeling and wiping failures.}
Figure~\ref{fig:rollout_failures_peel_wipe} separates failures during
contact from failures in acquisition and stage transitions. Producing a
strip or retaining a sponge does not ensure completion of the task.

\begin{figure}[!htbp]
    \centering
    \includegraphics[width=\linewidth]{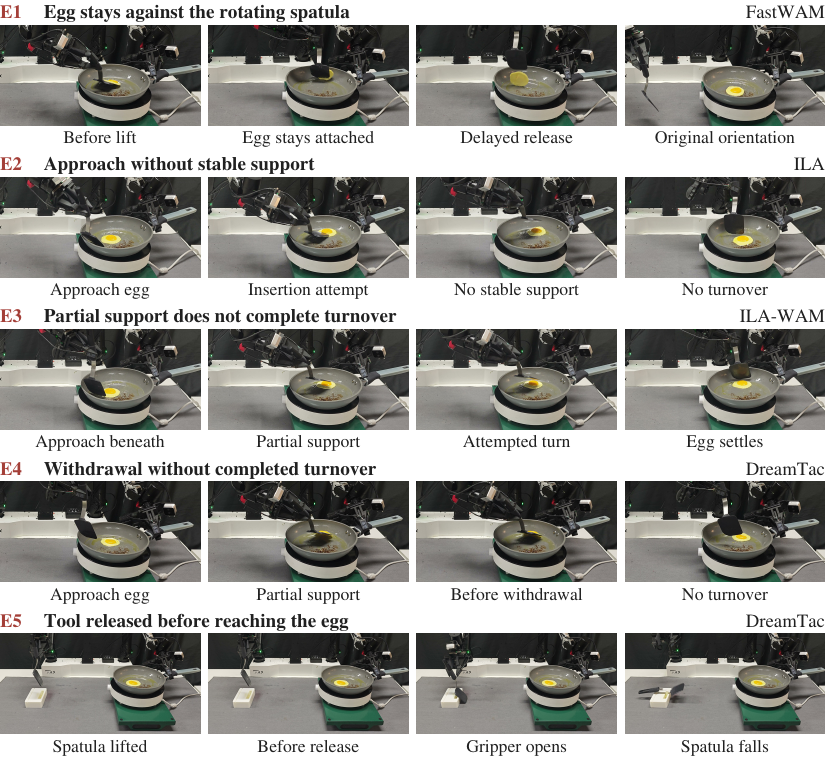}
    \caption{\textbf{Failure cases in egg flipping.}
    The rows distinguish attachment during rotation (E1), missing or
    inadequate support (E2--E3), withdrawal without turnover (E4),
    and premature tool release (E5). Frames retain the observed tool
    and object states; they do not isolate insertion depth or lifting
    clearance as a unique cause of failure.}
    \label{fig:rollout_failures_egg}
\end{figure}

\clearpage
\paragraph{Egg flipping: support, turnover, and tool retention.}
The five cases in Fig.~\ref{fig:rollout_failures_egg} fail at different
stages. In E1, the egg remains against the rotating spatula and returns
to its original orientation. E2 approaches without establishing stable
support; E3 obtains partial support but does not complete turnover.
E4 withdraws without flipping the egg. E5 instead releases the tool
before reaching the egg, preventing subsequent manipulation. Reaching
the target and rotating the spatula therefore do not, by themselves,
establish successful turnover.

\paragraph{Why the interaction phase matters.}
Contact must be established before lifting, maintained while the object
is supported, and released at the appropriate point in the turn. These
requirements differ even when the tool follows similar motions.
The cases motivate evaluating the local interaction state together
with temporal progress, rather than inferring completion from tool
position alone.

\begin{figure}[!htbp]
    \centering
    \includegraphics[width=\linewidth]{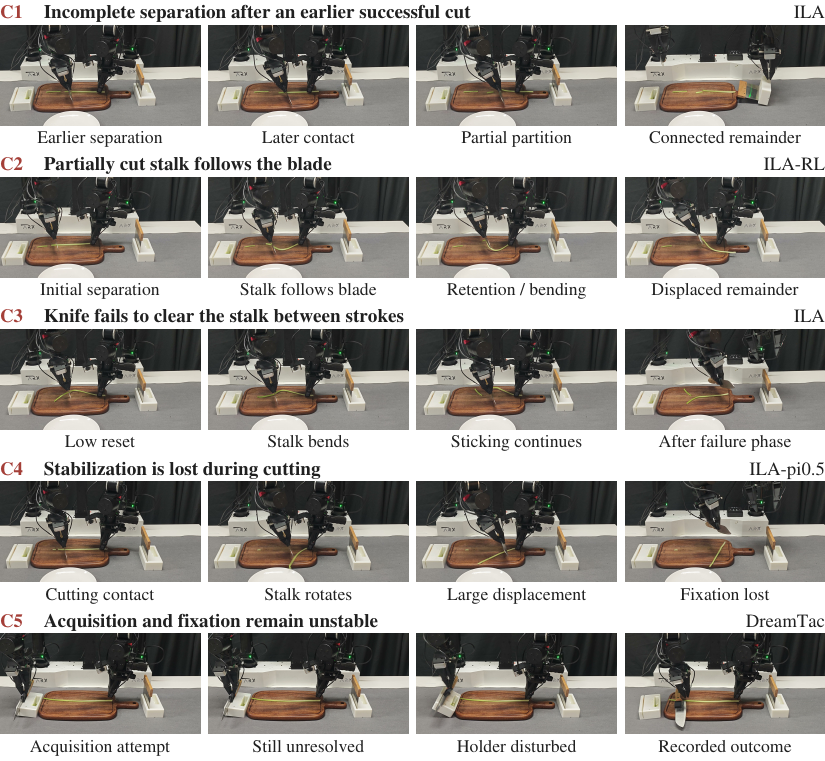}
    \caption{\textbf{Failure cases in celery cutting.}
C1--C5 illustrate incomplete separation, inadequate blade clearance,
loss of stabilization, and unstable acquisition.
A cut succeeds only if the celery is completely severed at the cutting
location before the knife clears the stalk.
Residual connections, including cases where the partially cut stalk
follows the lifted knife, count as failures.
Rows show selected failure phases; individual cuts within the same
recording may have different outcomes.}
    \label{fig:rollout_failures_cut}
\end{figure}

\clearpage
\paragraph{Celery cutting: local outcomes within a longer execution.}
Figure~\ref{fig:rollout_failures_cut} distinguishes incomplete separation
(C1), a partially cut stalk following the blade (C2), inadequate knife
clearance (C3), loss of stabilization (C4), and unresolved acquisition
and fixation (C5). Some recordings contain both successful and failed
cuts: C1 includes an earlier separated piece before a later incomplete
partition. We therefore assess individual cutting events. A subsequent
scooping or transfer motion does not establish that every preceding
cut succeeded.

\paragraph{Implications for the interaction latent $z$.}
Across tasks, establishing, maintaining, and releasing contact must be
coordinated with motion and changes in object state. Acquisition errors
can also propagate into later stages. These dependencies motivate
representing motor behavior together with tactile evolution in $z$,
and evaluating interactions locally within a rollout. The cases provide
behavioral context for the latent-space and reward analyses; they do
not identify a failure with a particular latent direction. The selected
examples illustrate distinct mechanisms of task failure, without
estimating their frequency or establishing a ranking between models.

\clearpage
\section{AI Use Statement}
\label{app:ai_use_statement}

Generative AI tools were used to polish the writing throughout the manuscript and assist with the selection of figure color palettes. All AI-assisted revisions and color choices were reviewed and approved by the authors. The authors take full responsibility for the accuracy and integrity of the final submission, including its technical descriptions, citations, experimental claims, and figures.

\end{document}